\documentclass{article}

\usepackage[numbers]{natbib}
\usepackage[preprint]{neurips_2022}

\usepackage[dvipsnames]{xcolor}         
\definecolor{linkColor}{rgb}{0.18,0.39,0.62}
\usepackage[utf8]{inputenc} 
\usepackage[T1]{fontenc}    
\usepackage[colorlinks=true,linkcolor=linkColor,citecolor=linkColor,filecolor=linkColor,urlcolor=linkColor]{hyperref}       
\usepackage{url}            
\usepackage{booktabs}       
\usepackage{amsfonts}       
\usepackage{nicefrac}       
\usepackage{microtype}      

\usepackage{graphicx}
\usepackage{arydshln}
\usepackage{booktabs}
\usepackage{multirow}
\usepackage{caption}
\usepackage{subcaption}
\usepackage{makecell}
\usepackage{csquotes}
\usepackage{epigraph}
\usepackage{amsmath}
\usepackage{xcolor}  
\usepackage{cleveref}
\usepackage{pifont}
\usepackage{listings}
\usepackage{amssymb}

\RequirePackage{algorithm}
\RequirePackage{algorithmic}

\usepackage{multirow}
\usepackage{amsmath}
\usepackage{capt-of}
\usepackage{tabularx}
\usepackage{epsfig}
\usepackage{amssymb}
\usepackage{amsfonts}
\usepackage{booktabs}
\usepackage{scalerel}
\usepackage[inline]{enumitem}
\usepackage{listings}
\usepackage{varwidth}
\usepackage[export]{adjustbox}
\usepackage{tikz}
\usetikzlibrary{tikzmark}

\usepackage{stmaryrd}
\usepackage{bbm}
\usepackage{wrapfig}
\usepackage{pifont}

\definecolor{deepblue}{rgb}{0,0,0.5}
\definecolor{officeblue}{RGB}{0,102,204}
\definecolor{deepred}{rgb}{0.6,0,0}
\definecolor{deepgreen}{rgb}{0,0.5,0}
\definecolor{mybrickred}{RGB}{182,50,28}

\definecolor{fillcolor}{RGB}{216,217,252}

\usepackage{etoolbox}
\usepackage{framed}

\newif\ifxetexorluatex
\ifxetex
  \xetexorluatextrue
\else
  \ifluatex
    \xetexorluatextrue
  \else
    \xetexorluatexfalse
  \fi
\fi
\newcommand*\quotesize{60} 
\newcommand*{\openquote}
   {\tikz[remember picture,overlay,xshift=-4ex,yshift=-2.5ex]
   \node (OQ) {\fontsize{\quotesize}{\quotesize}\selectfont``};\kern0pt}

\newcommand*{\closequote}[1]
  {\tikz[remember picture,overlay,xshift=4ex,yshift={#1}]
   \node (CQ) {\fontsize{\quotesize}{\quotesize}\selectfont''};}

\colorlet{shadecolor}{white}

\newcommand*\shadedauthorformat{\emph} 

\newcommand*\authoralign[1]{%
  \if#1l
    \def\authorfill{}\def\quotefill{\hfill}
  \else
    \if#1r
      \def\authorfill{\hfill}\def\quotefill{}
    \else
      \if#1c
        \gdef\authorfill{\hfill}\def\quotefill{\hfill}
      \else\typeout{Invalid option}
      \fi
    \fi
  \fi}
{\authoralign{#1}
\ifblank{#2}
   {\def\shadequoteauthor{}\def\yshift{-2ex}\def\quotefill{\hfill}}
   {\def\shadequoteauthor{\par\authorfill\shadedauthorformat{#2}}\def\yshift{2ex}}
\begin{snugshade}\begin{quote}\openquote}
{\shadequoteauthor\quotefill\closequote{\yshift}\end{quote}\end{snugshade}}

\usepackage{amsmath,amsfonts,bm}

\def\eqref#1{equation~\ref{#1}}

\def\1{\bm{1}}

\DeclareMathAlphabet{\mathsfit}{\encodingdefault}{\sfdefault}{m}{sl}
\SetMathAlphabet{\mathsfit}{bold}{\encodingdefault}{\sfdefault}{bx}{n}

\newcommand\our{\text{Q-Sparse}}
\newcommand\llama{\text{LLaMA LLM}}

\title{\our{}: All Large Language Models can be \\ \colorbox{gray!30}{Fully} Sparsely-Activated}

\vspace{-2.8cm}
 \author{
 Hongyu Wang\thanks{~Equal contribution. $\diamond$ Corresponding author. S. Ma, F. Wei are with Microsoft Research. H. Wang and R. Wang are with University of Chinese Academy of Sciences.}~~~~Shuming Ma\footnotemark[1]~~~~Ruiping Wang~~~~Furu Wei$^{\diamond}$ \\
 {\href{https://aka.ms/GeneralAI}{https://aka.ms/GeneralAI}}
\vspace{-0.4cm}
 \\}

\begin{document}
\maketitle
\vspace{-0.5cm}
\begin{abstract}
\vspace{-0.3cm}

We introduce, \textbf{\our{}}, a simple yet effective approach to training sparsely-activated large language models (LLMs). \our{} enables \textbf{full sparsity of activations} in LLMs which can bring significant efficiency gains in inference. This is achieved by applying top-$K$ sparsification to the activations and the straight-through-estimator to the training. We also introduce \textbf{Block \our{}} for batch training and inference. The key results from this work are, (1) \our{} can achieve results comparable to those of baseline LLMs while being much more efficient at inference time;
(2) We present an inference-optimal scaling law for sparsely-activated LLMs; (3) \our{} is effective in different settings, including training-from-scratch, continue-training of off-the-shelf LLMs, and finetuning; (4) \our{} works for both full-precision and 1-bit LLMs (e.g., \textbf{BitNet b1.58}~\cite{bitnet}). Particularly, the synergy of BitNet b1.58 and \our{} (can be equipped with MoE) provides the cornerstone and a clear path to revolutionize the efficiency, including cost and energy consumption, of future LLMs.

\vspace{-0.3cm}

\begin{figure}[h]
    \centering
    \begin{subfigure}{0.4\textwidth}
        \centering
        \includegraphics[width=\linewidth]{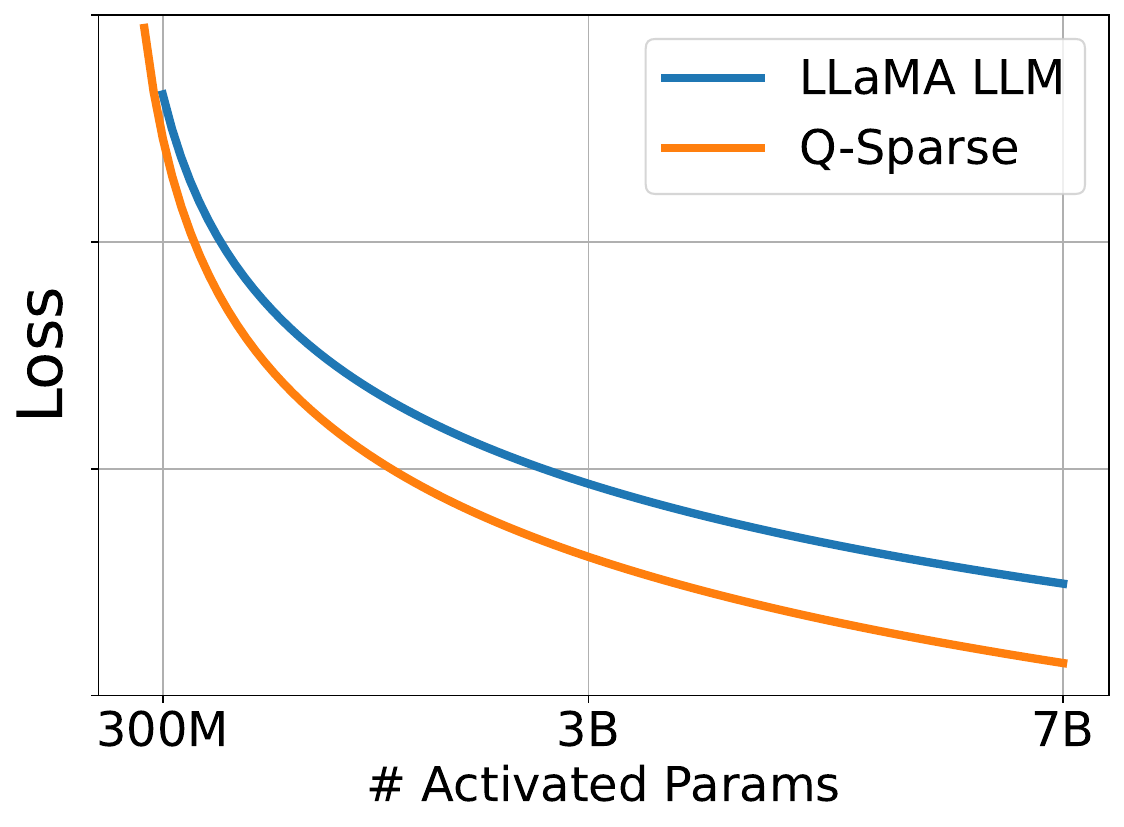}
    \end{subfigure}
    \begin{subfigure}{0.4\textwidth}
        \centering
        \includegraphics[width=\linewidth]{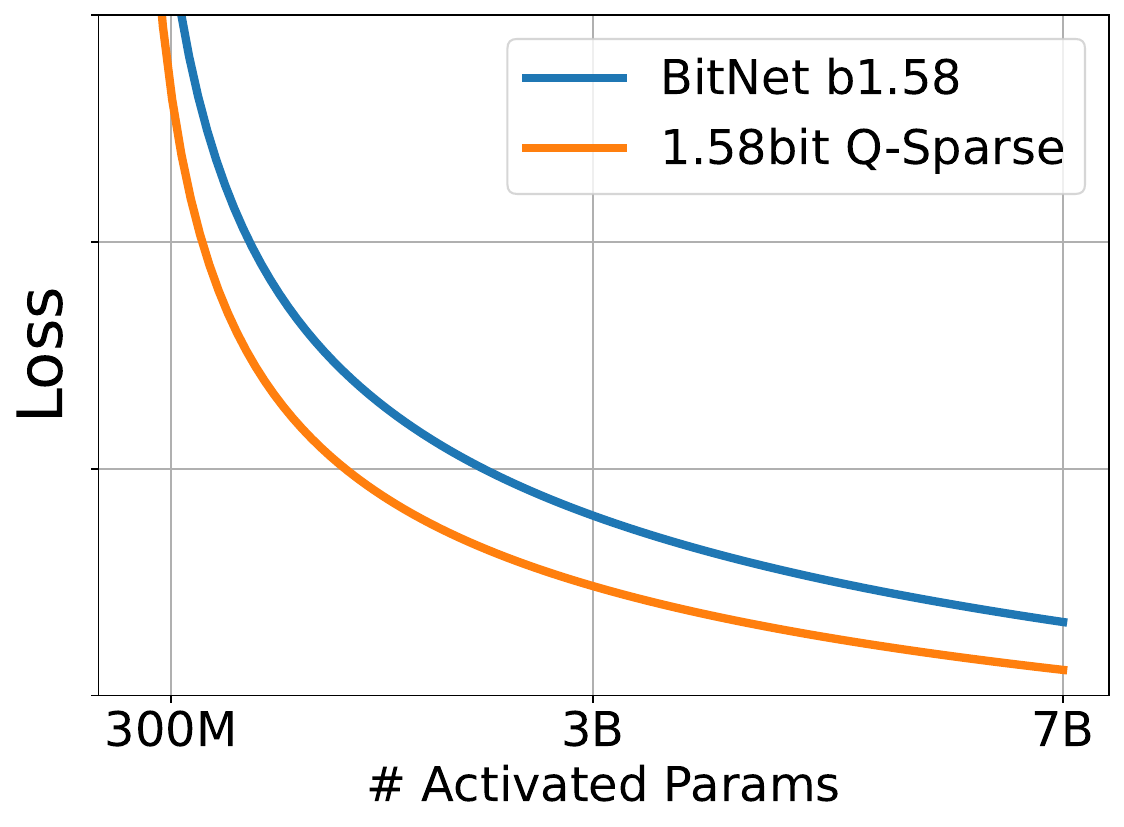}
    \end{subfigure}
    \includegraphics[width=0.7\linewidth]{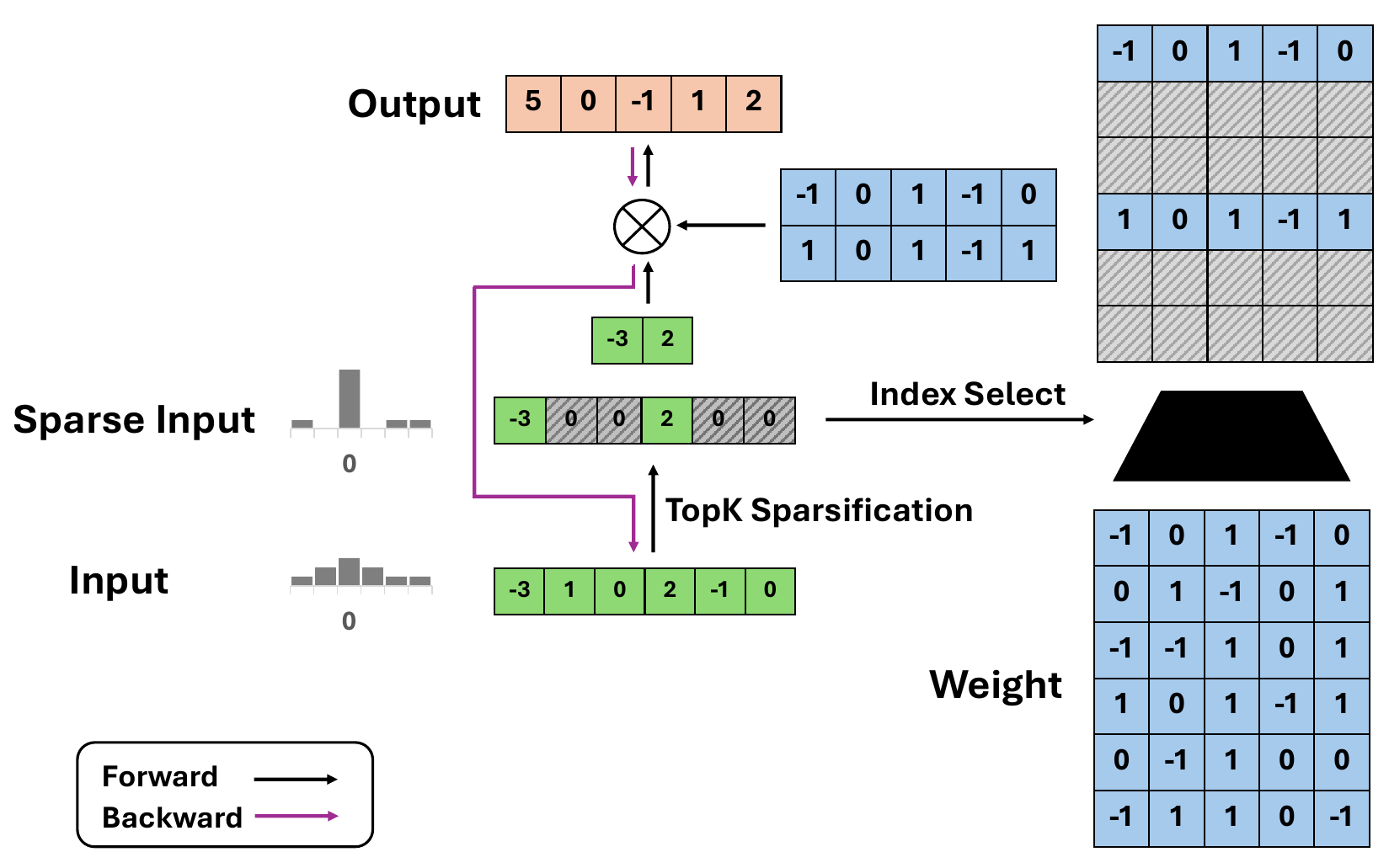}
    \vspace{-0.25cm}
    \caption{\our{} achieves a superior inference-optimal scaling law than the dense models. It saves significant compute of matrix multiplication by top-$K$ sparsification of the activations.}
    \label{fig1}
\end{figure}

\end{abstract}

\newpage
\section{Fully Sparsely-Activated LLMs}

Large language models (LLMs) have achieved remarkable performance on a wide range of natural language processing (NLP) tasks. However, the deployment of LLMs in real-world applications is challenging due to their high computational cost and memory footprint, especially during the inference stage. To address this challenge, recent works~\cite{bitnetx, bitnet, turbosparse, sheared, specdecoding} have focused on improving the efficiency of LLMs with various approaches, including quantization~\cite{bitnetx, bitnet, gptq}, pruning~\cite{sheared}, distillation~\cite{minillm}, better decoding~\cite{specdecoding}, and so on. One promising approach is to use sparsity to reduce the number of activated parameters in LLMs.

Sparsity contributes two factors to the efficiency of LLMs. First, sparsity can reduce the amount of computation of the matrix multiplication as zero elements are not computed. Second, sparsity can reduce the amount of input/output (I/O) that transfers the parameters between the memory and the computation units. The I/O transfer serves as the major bottleneck in the inference stage of LLMs.

One common approach to sparsity in LLMs is to use weight sparsity, which prunes the model weights to save the computation. However, unstructured weight sparsity is difficult to parallelize in GPU devices, while structured weight sparsity has a large impact to the accuracy of the model. 

Another approach is to use activation sparsity, which reduces the number of activated elements in the activation tensors. Activation sparsity can be achieved by using the mixture-of-experts (MoE) mechanism~\cite{gshard,switch}, modifying the activation function~\cite{relu_back,turbosparse}, or predicting the position to be sparsed~\cite{dejavu}. However, these approaches do not enable full sparsity of activations in LLMs, which can limit the efficiency gains during the inference stage. Moreover, compared to the dense models, the scaling laws for the sparsely-activated LLMs have not been well studied.

To explore the full potential of sparsity in LLMs, we introduce \textbf{\our{}}, a simple yet effective approach to enable full sparsity of activations in LLMs. The major modification on LLMs is in the linear projection (i.e., matrix multiplication). As shown in Figure~\ref{fig1}, for each linear projection, it has a top-K sparsification function that selects the top-K activations in the input tensor. For the backprogation, we use the straight through estimator to compute the gradients of the activations. We also introduce a squared ReLU function for the feed-forward layers to further improve the sparsity of the activations. \our{} can be used with both full-precision and quantized LLMs. Furthermore, we introduce, \textbf{Block \our{}}, a block sparsity implementation to make \our{} compatible with batch training and inference.

To study the scaling law of sparsely-activated LLMs, we conduct a series of scaling experiments and derive an inference-optimal scaling law for sparsely-activated LLMs. We summarize the findings from the scaling experiments and the implications of the scaling law as below:

\begin{itemize}
    \item The performance of the sparsely-activated models is better than the dense baselines with the same inference compute budget (i.e., activated parameters or FLOPs).
    \item As the parameters $N$ scales, the performance gap between the sparsely-activated models and the dense baselines decreases.
    \item The performance of the sparsely-activated models with around 40\% sparsity ratio can match the performance of the dense baselines with the same model size and training tokens.
    \item Given the same inference budget $N_a$, a sparsely-activated full-precision model with a sparsity ratio of 45.58\% (or $1.84N_{a}$ parameters) can achieve the best performance. For the 1.58-bit models, the optimal sparsity ratio is 61.25\%.
\end{itemize}

We also conduct experiments to evaluate the effectiveness of \our{} in different settings, including training-from-scratch, continue-training of off-the-shelf LLMs, and finetuning. We show that \our{} can achieve results comparable to those of baseline LLMs with the same training cost while being much more efficient at inference time.

\section{\our{}}

\subsection{Architecture}

The \our{} architecture is based on the Transformer architecture~\cite{transformer, llama} with modifications to enable sparsity in the activations. 

\textbf{Top-K Sparsity}

The Transformer architecture uses \emph{nn.Linear} to perform the projection in both attention and feed-forward layers, which can be written as:
\begin{equation}
    \mathbf{Y} = \mathbf{X} \cdot \mathbf{W}^T
\end{equation}

where $\mathbf{X} \in \mathbb{R}^{N \times D}$ is the input tensor, $\mathbf{W} \in \mathbb{R}^{M \times D}$ is the weight tensor, and $\mathbf{Y} \in \mathbb{R}^{N \times M}$ is the output tensor. The \emph{nn.Linear} operation is equivalent to the matrix multiplication operation.

We introduce a top-K sparsity function on top of the matrix multiplication operation. The top-K sparsity function is defined as:

\begin{equation}
    \mathbf{Y} = (\mathbf{X} \odot \mathbf{M}) \cdot \mathbf{W}^T
\end{equation}

\begin{equation}
    \mathbf{M} = \text{Top}_k (\mathbf{|X|})
\end{equation}

where $\mathbf{M} \in \mathbb{R}^{N \times D}$ is the mask tensor that indicates the top-K activations in the input tensor $\mathbf{X}$ in terms of the absolute values, $\odot$ is the element-wise multiplication operation, and $\text{Top}_k$ is the function that selects the top-K elements in the tensors.

To reduce the interval around zero, we re-scale the tensor by its $L_2$ norm after performing the top-K sparsity function.

\textbf{Quantized Top-K Sparsity}

Recent works~\cite{bitnet} have shown that quantization can be used to reduce the memory footprint and computational cost of LLMs without the loss of performance. We introduce a quantized version of the top-K sparsity function. The quantized top-K sparsity function is defined as:
\begin{equation}
    \mathbf{Y} = (\text{Q}(\mathbf{X}) \odot \mathbf{M}) \cdot \mathbf{W}^T
\end{equation}

where $\text{Q}(\cdot)$ is the quantization function that quantizes the input tensor $\mathbf{X}$ to a 8-bit representation:
\begin{equation}
    \text{Q}(X) = \text{RoundClip}(\frac{127}{\gamma+\epsilon}\mathbf{X}, -128, 127)
\end{equation}

\begin{equation}
    \gamma = \max(|\mathbf{X}|)
\end{equation}

\begin{equation}
    \text{RoundClip}(X, a, b) = \min(\max(\text{round}(X), a), b)
\end{equation}

where $\epsilon$ is a small constant to avoid division by zero, and $\gamma$ is the maximum absolute value in the input tensor $\mathbf{X}$.

\our{} can be used with both full-precision and quantized LLMs. Specifically, the quantized version of \our{} is compatible with 1-bit LLMs, such as BitNet b1.58~\cite{bitnet}. When using \our{} with 1-bit LLMs, the quantization function is performed on the weight tensor $\mathbf{W}$:
\begin{equation}
    \mathbf{Y} = (\text{Q}(\mathbf{X}) \odot \mathbf{M}) \cdot \text{Q}_{w}(\mathbf{W})^T
\end{equation}

where $\text{Q}_{w}(\cdot)$ is the quantization function that quantizes the weight tensor $\mathbf{W}$ to a 1.58-bit representation:
\begin{equation}
    \text{Q}_{w}(W) = \text{RoundClip}(\frac{\mathbf{W}}{\alpha+\epsilon}, -1, 1)
\end{equation}

where $\alpha$ is the mean absolute value in the weight tensor $\mathbf{W}$:
\begin{equation}
    \alpha = \text{mean}(|\mathbf{W}|)
\end{equation}

\textbf{Squared ReLU}

To further improve the sparsity of the activations, we use the squared ReLU function~\cite{primer} for the feed-forward layers. The squared ReLU function is defined as $\text{ReLU}(\mathbf{X})^2$.

Following the LLaMA architecture, we use the gated linear unit (GLU) for the feed-forward layers. The squared ReLU function is applied with the GLU function into a ReLU$^2$GLU function. The ReLU$^2$GLU function is defined as:

\begin{equation}
    \text{ReLU}^2\text{GLU}(\mathbf{X}) = \mathbf{X} \mathbf{W}_{\text{up}}^{T} \odot \text{ReLU}^2(\mathbf{X} \mathbf{W}_{\text{gate}}^T)
\end{equation}

\textbf{Block \our{}}

While the top-k sparsification can be used in the single-sample mode, it is not friendly with the batch mode for the current GPU devices. Recent work~\cite{nmfinegrain,sparsenm} shows that N:M sparsity, where N out of M consecutive elements to be zero, is more hardware friendly and can be used in the batch mode with an optimized GPU kernel.
To leverage this feature of the modern GPU devices, we introduce Block \our{}. The key idea of Block \our{} is to apply the top-K sparsity function on the activations in the block level, and the block size is set to $M$ so that there are always $M-K$ zeros out of $M$ consecutive values. The top-K sparsity function is applied to the activations in each block independently. The block level sparsity can be used to reduce the memory footprint and computational cost of the LLMs in the batch mode.

\subsection{Training}

Most of the existing works~\cite{relu_back} on training sparsely-activated models use the vanilla back-propagation algorithm to compute the gradient through the sparsity function:
\begin{equation}
    \frac{\partial \mathbf{Y}}{\partial \mathbf{X}} = \frac{\partial \mathbf{Y}}{\partial (\mathbf{X} \odot \mathbf{M})} \odot \mathbf{M}
\end{equation}
where $\mathbf{M}$ is the mask tensor that indicates the top-K activations in the input tensor $\mathbf{X}$, and $\odot$ is the element-wise multiplication operation. 

The vanilla back-propagation algorithm has a limitation. It zero-outs the gradients of the non-activated elements, which can lead to the vanishing gradient problem, especially when the sparsity ratio is high. In this work, we propose to use the straight-through estimator~\cite{ste} to back-propagate the gradients through the sparsity function. In this way, the gradients are passed through the sparsity function without being zeroed-out. The straight-through estimator is defined as:

\begin{equation}
    \frac{\partial \mathbf{Y}}{\partial \mathbf{X}} = \frac{\partial \mathbf{Y}}{\partial (\mathbf{X} \odot \mathbf{M})}
\end{equation}

\begin{figure}[t]
    \centering
    \includegraphics[width=\textwidth]{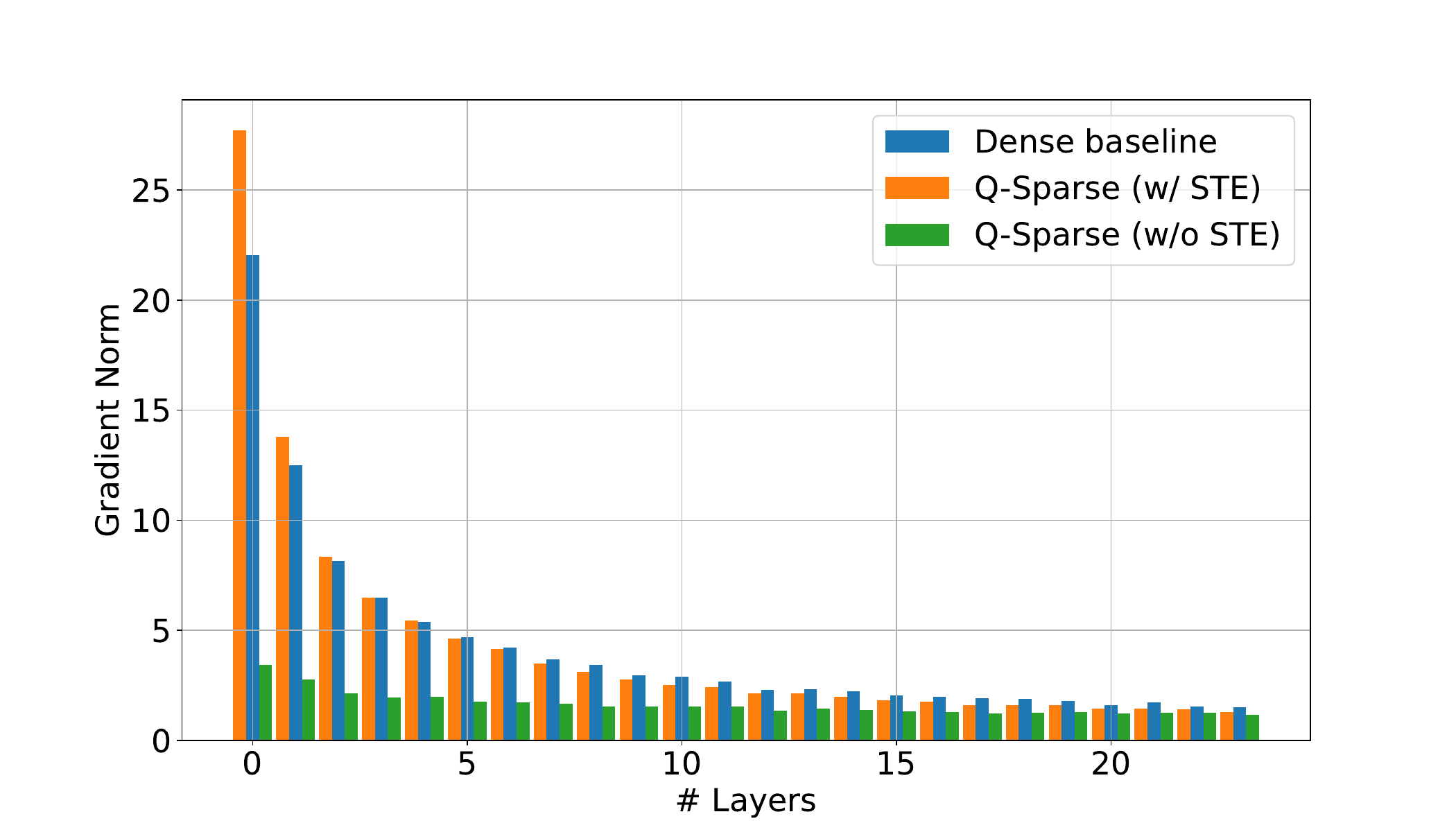}
    \caption{The average magnitude of each projection's gradient of dense baseline, \our{} with and without STE across different layers. The visualization is conducted with 300M model size on a subset of the valid set of C4~\cite{c4}. It shows that the gradient vanishes without STE.}
    \label{fig:grad}
\end{figure}

We visualize the average $l2$ norm of each projection's gradient across different layers for dense model, \our{} with and without STE. We adopt top-K as 50\% for \our{}. Without STE, the gradient is much smaller at the bottom layers, while STE can preserve the magnitude of the gradients. As shown in Figure~\ref{fig:grad}, STE estimator significantly eases the issue of gradient vanishing, especially at the bottom of the layers. We present more visualizations for each components in the Appendix~\ref{ap:vis}.

\subsection{\our{} for Continue-Train and Finetuning Settings}

\our{} can be used in different settings, including training-from-scratch, continue-training, and finetuning. In the continue-train and finetuning settings, we use the same architecture and training procedure as in the training-from-scratch setting. The only difference is that we initialize the model with the pre-trained weights and continue training with the sparsity function enabled.

For the pre-trained models that do not have the squared ReLU function in the feed-forward layers, we apply the top-K sparsity function after the activated function (e.g., SiLU) in the feed-forward layers. It can improve the sparsity of the activations without changing the model architecture.

\section{Scaling Laws}

Recent work on large language models has shown that the performance of LLMs scales with the model size and the amount of training data. \cite{chinchilla} argues that the converged performance of a dense Transformer model with $N$ parameters follows a power-law scaling law, which can be written as:
\begin{equation}
    L(N) \triangleq E + \frac{A}{N^{\alpha}}
\end{equation}

where $L(N)$ is the performance of the model with $N$ parameters, $E$ is the performance of the model with infinite parameters, $A$ is a constant, and $\alpha$ is the scaling exponent. Note that the number of training tokens are fixed in this setting, which is part of the constant $E$.

In this work, we investigate the scaling law of sparsely-activated LLMs. We find that the performance of sparsely-activated LLMs also follows a power-law scaling law, which can be written as:
\begin{equation}
    L(N, S) \triangleq E + \frac{A(S)}{N^{\alpha}}
\end{equation}

\begin{equation}
    A(S) = B + C \exp{(\frac{\beta}{1-S})}
\end{equation}

where $L(N, S)$ is the performance of the sparsely-activated model with $N$ parameters and a sparsity ratio of $S$, and $\alpha$ and $\beta$ are the scaling exponents.

In the following part, we will introduce how we derive the scaling law and the corresponding findings.

\subsection{Scaling Experiments and Findings}

To determine the form of the scaling law of sparse-activated LLMs, we begin with a series of scaling experiments. In the experiments, we train a series of language models with \our{} of various scales, ranging from 300M to 7B. The models are trained on the Redpajama dataset~\cite{redpajama}. We use the Sentencepiece tokenizer from LLaMA to preprocess data. Besides \our{}, we also train the dense baselines with the same datasets and settings. More details can be found in the Appendix~\ref{ap:hyper}.

\begin{figure}[t]
    \centering
    \begin{subfigure}{0.49\textwidth}
        \centering
        \includegraphics[width=\linewidth]{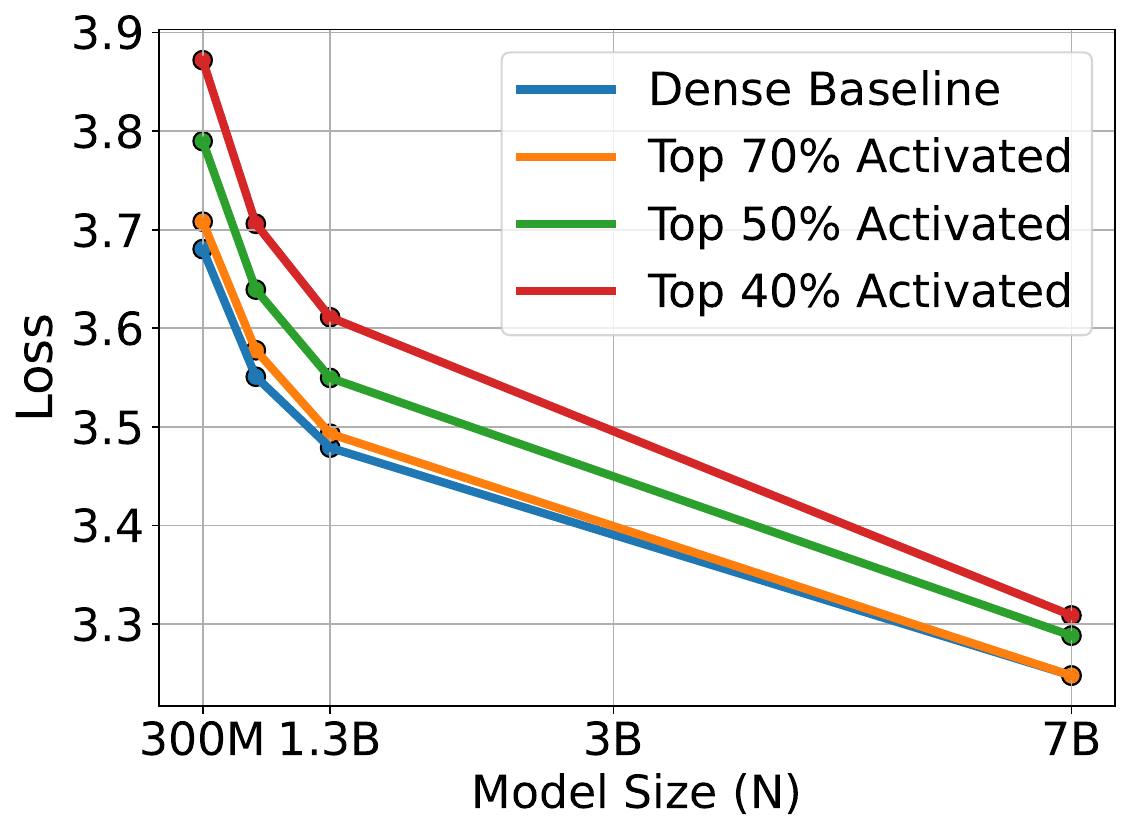}
    \end{subfigure}
    \begin{subfigure}{0.49\textwidth}
        \centering
        \includegraphics[width=\linewidth]{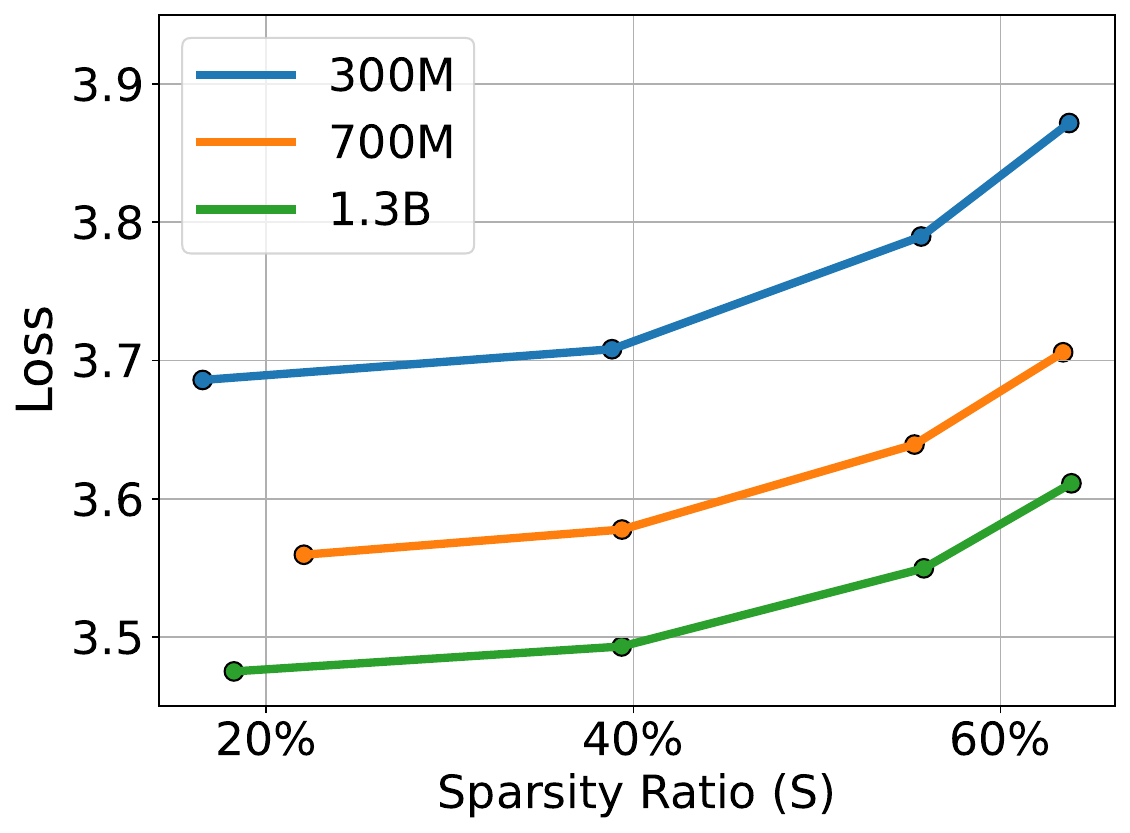}
    \end{subfigure}
    \caption{The scaling curves of the sparsely-activated models regrading to the model size given a fixed sparsity ratio $S$ (Left), and regrading to the sparsity ratio given a fixed model size $N$ (Right).}
    \label{fig:scaling}
\end{figure}

The observed losses of the sparsely-activated models and the dense baselines are shown in Figure~\ref{fig:scaling}. We summarize the findings as below:
\begin{itemize}
    \item The performance of the sparsely-activated models scales with the model size and the sparsity ratio.
    \item Given a fixed sparsity ratio $S$, the performance of the sparsely-activated models follows a power-law scaling law with regards to the model size $N$.
    \item Given a fixed parameters $N$, the performance of the sparsely-activated models follows an exponential-law scaling law with regards to the sparsity ratio $S$.
    \item As the parameters $N$ scales, the performance gap between the sparsely-activated models and the dense baselines decreases.
\end{itemize}

According to these findings, our main hypothesis is that the performance of the sparsely-activated models follows a combination of a power-law scaling law with regards to the model size $N$ and an exponential-law scaling law with regards to the sparsity ratio $S$.

\subsection{Power Law in the Model Size $N$}

With a fixed sparsity ratio $S$, the scaling law should follows \cite{scalinglawlm}'s scaling law, which can be written as:
\begin{equation}
    L(N, S) \triangleq E + \frac{A(S)}{N^{\alpha(S)}}
\end{equation}

where $\alpha(S)$ is the scaling exponent, and the scaling factor $A(S)$ is a function of the sparsity ratio $S$. Given any model size $N$, the function $L(N, S)$ should follow the Lipschitz continuity with regards to the sparsity ratio $S$. Therefore, the scaling exponent $\alpha(S)$ should be a non-decreasing function. Given any model size $N$, the function $L(N, S)$ is increasing with the sparsity ratio $S$, so $\alpha(S)$ should be a non-increasing function. Above all, the scaling exponent $\alpha(S)$ should be a constant, and the scaling function can be written as:
\begin{equation}
    L(N, S) \triangleq E + \frac{A(S)}{N^{\alpha}}
\end{equation}

\begin{figure}[t]
\centering
\begin{subfigure}{0.38\textwidth}
    \centering
    \includegraphics[width=\linewidth]{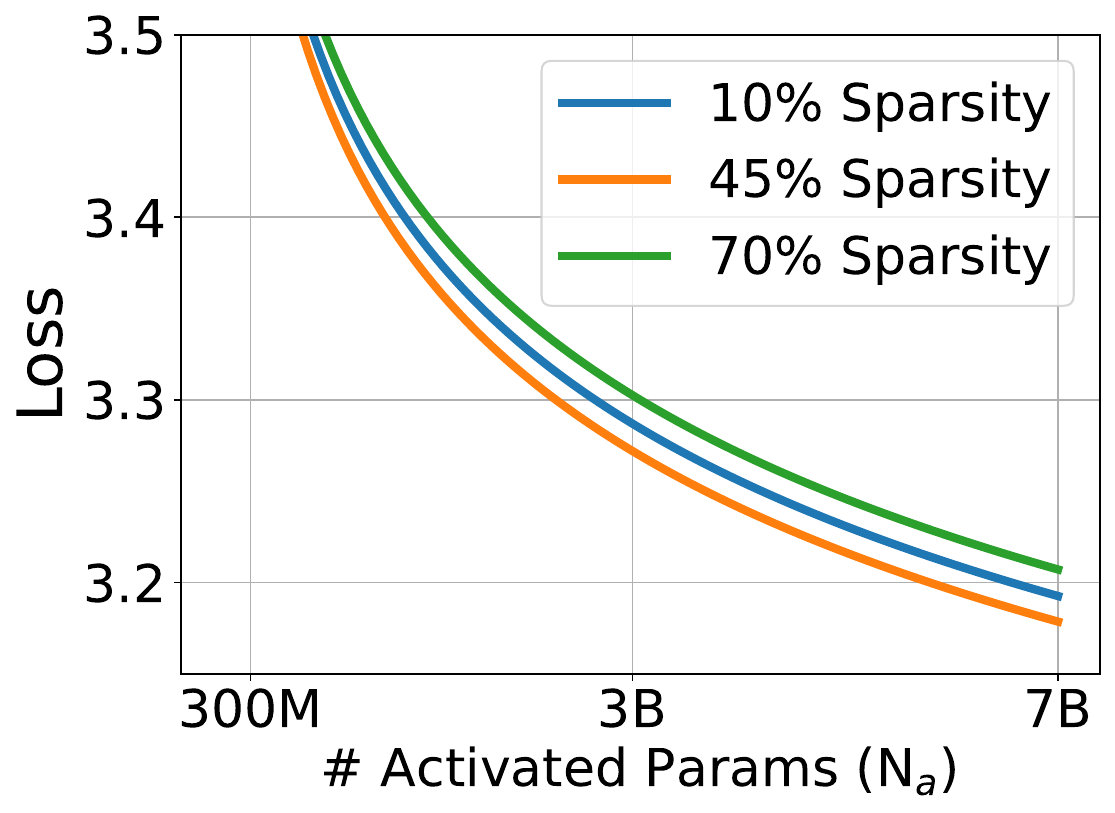}
\end{subfigure}
\begin{subfigure}{0.61\textwidth}
    \centering
    \includegraphics[width=\linewidth]{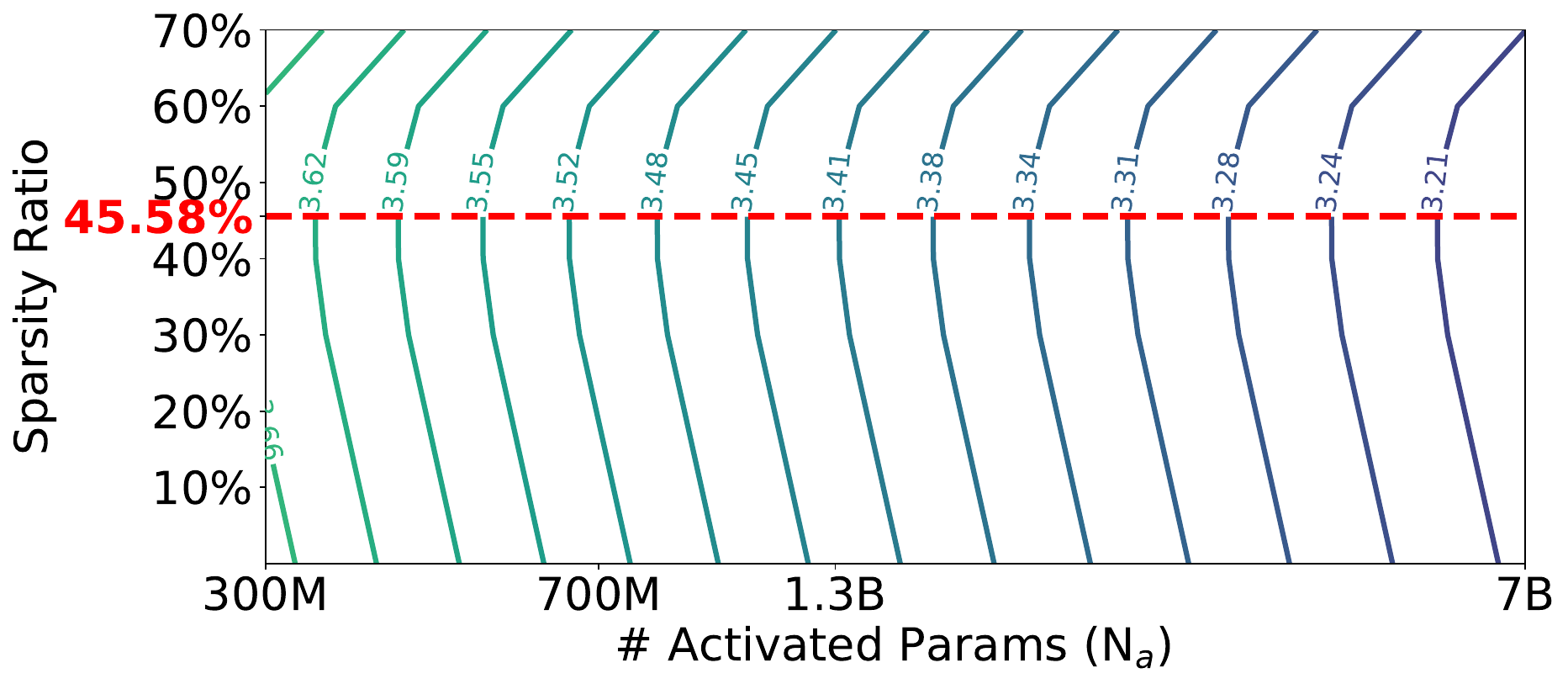}
\end{subfigure}
\begin{subfigure}{0.38\textwidth}
    \centering
    \includegraphics[width=\linewidth]{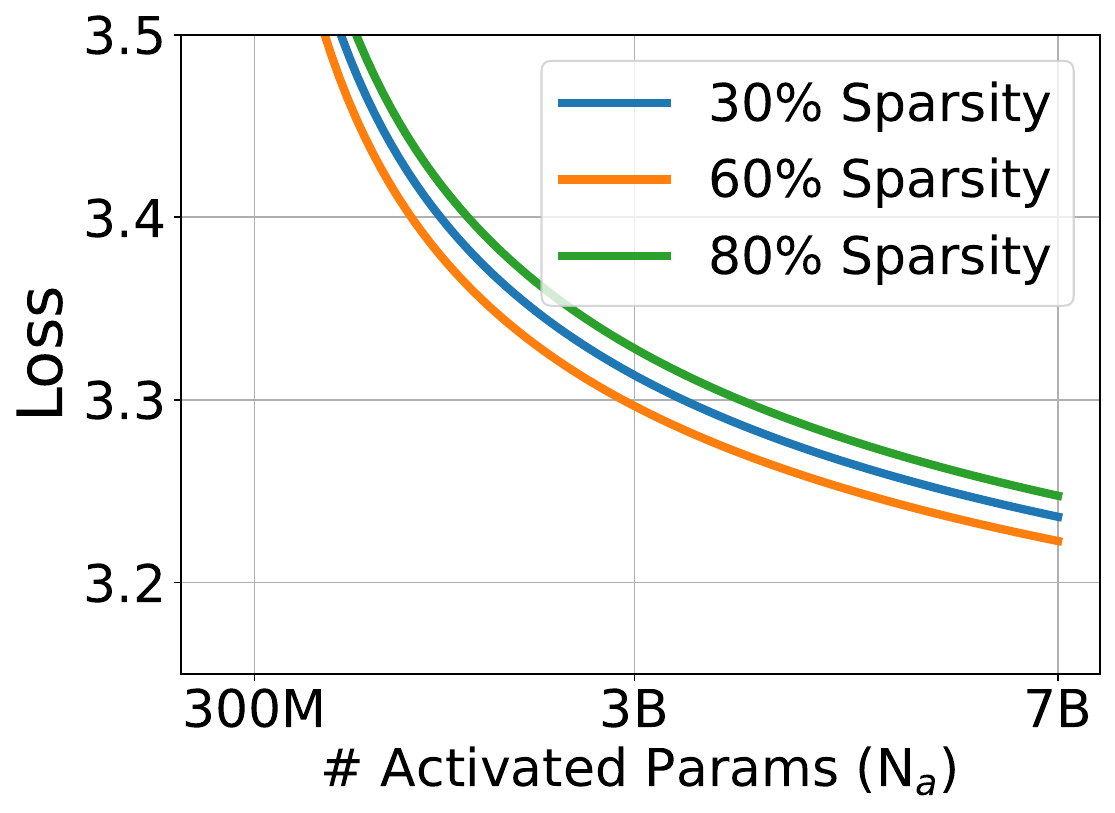}
\end{subfigure}
\begin{subfigure}{0.61\textwidth}
    \centering
    \includegraphics[width=\linewidth]{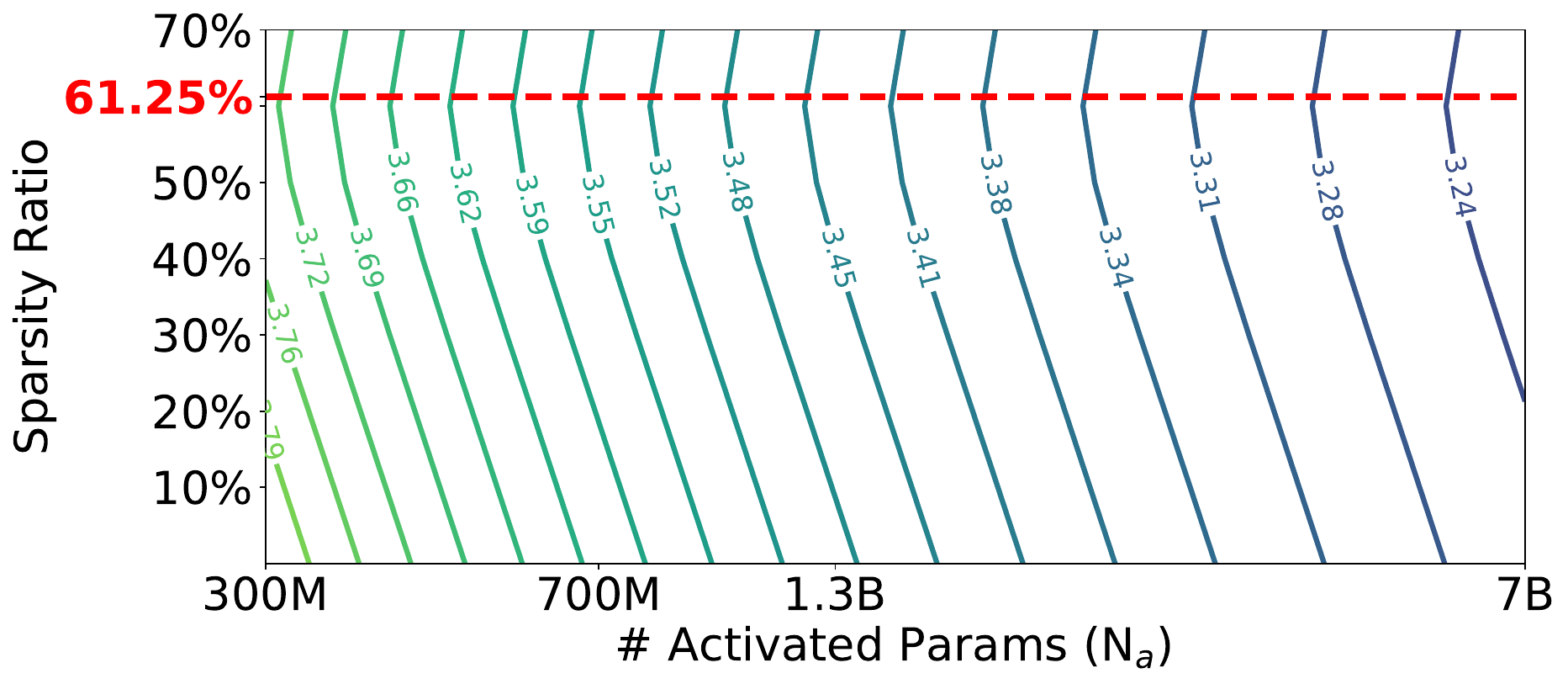}
\end{subfigure}
\caption{The inference-optimal scaling curves of the sparsely-activated models with full-precision (Top) and 1.58-bit (Bottom) weight. It shows that a sparisty of 45.58\% for full-precision models and 61.25\% for 1.58-bit models can achieve the best performance with the same inference compute budget (i.e., activated parameters or FLOPs).}\label{fig:inference_optimal}
\end{figure}

\subsection{Exponential Law in the Sparsity Ratio $S$}

According to the above finding, the performance of the sparsely-activated models follows an exponential-law scaling law with regards to the sparsity ratio $S$. Therefore, the scaling factor $A(S)$ should also follow an exponential law. Besides, given any model size $N$, the scaling function is increasing with the sparsity ratio $S$. Therefore, the scaling factor $A(S)$ should be a non-decreasing function. The scaling factor $A(S)$ can be written as:
\begin{equation}
    A(S) = B + C \exp{(\frac{\beta}{1-S})}
\end{equation}

where $B$ is the scaling factor for extremely sparse LLMs, $C$ is the scaling factor for dense LLMs, and $\beta$ is the scaling exponent of the scaling factor $A(S)$ with regards to the sparsity ratio $S$.

\subsection{Fitting the Parameters}


We fit the parameters of the scaling law to the observed losses of the sparsely-activated models. We use the L-BFGS algorithm~\cite{lbfgs} to minimize the Huber loss~\cite{huber} between the predicted and observed log loss.

\begin{equation}
    \min_{E, B, C, \beta, \alpha} \sum_{\text{Runs } i} \text{Huber}_{\delta} \left( \log \hat{L}(N_i, S_i) - \log L_i \right)
\end{equation}

Following~\cite{chinchilla}, $\delta$ is set as $10^{-3}$. We select the best fit from a grid of initialisations around possible local optimas. $E$, $B$, $C$, $\alpha$ and $\beta$ are estimated as 1.86, 0.01, 1.89, 0.10 and 0.05, respectively.

\subsection{Diminishing Gap between Sparsely-Activated Models and Dense Baselines}

Given the above scaling law, we can derive the performance of the sparsely-activated models and the dense baselines with the same model size $N$ and the same sparsity ratio $S$. The performance gap between the sparsely-activated models and the dense baselines decreases as the model size $N$ scales. The performance gap can be written as:
\begin{align}
    L(N, S) - L(N, 0) & = \frac{A(S)}{N^{\alpha(S)}} - \frac{A(0)}{N^{\alpha(0)}} \\
    & = \frac{A(0)}{N^{\alpha}}(\frac{A(S)}{A(0)} - 1)
\end{align}

Since $\alpha$ is a constant that satisfies $\alpha > 0$, the performance gap decreases as the model size $N$ scales. It means that given a large enough model size $N$, the performance of the sparsely-activated models can eventually match the performance of the dense baselines with the same model size.

\subsection{Inference-Optimal Scaling Law}

The scaling law can also be transformed into a form that is dependent on the activated parameters $N_a$, which reflects the effective compute (i.e., FLOPs) of the model during inference:
\begin{equation}
    L(N_a, S) \triangleq E + A(S)(\frac{1-S}{N_a})^{\alpha}
\end{equation}

where $N_a$ is the number of activated parameters in the model, which is equal to $N \times (1-S)$. Since $A(S)$ is an increasing function and $(1-S)^\alpha$ is 
a decreasing function, there exists a sparsity ratio $S^* > 0$ that minimizes the loss of the sparsely-activated models. This leads to the inference-optimal scaling law of the sparsely-activated models:
\begin{equation}
    L(N_a) \triangleq E + A(S^*)(\frac{1-S^*}{N_a})^{\alpha}
\end{equation}

It shows that the performance of the sparsely-activated models is better than the dense baselines with the same inference compute budget. We further solve the optimal sparsity ratio $S^*$, finding that $S^* \approx 45.58\%$. It means that a sparsely-activated model with a sparsity ratio of 45.58\% (or $1.84N_{a}$ parameters) can achieve the best performance with the same inference budget $N_a$.
We follow the same process to estimate the inference-optimal scaling law for 1.58-bit \our{} models. We find that the optimal sparsity ratio is 61.25\% (or $2.58N_{a}$ parameters). Figure~\ref{fig:inference_optimal} shows the inference-optimal scaling curves of the sparsely-activated models with full-precision and 1.58-bit weight. It shows that with the same performance, the sparsely-activated models can achieve a significant reduction in the number of activated parameters or FLOPs during inference.

The inference-optimal scaling law shows that the performance of the sparsely-activated models can be optimized by adjusting the sparsity ratio $S$. It can be used to guide the training of the sparsely-activated models and to optimize the performance of the models during inference.

\begin{figure}[t]
    \centering
    \begin{subfigure}{0.48\textwidth}
        \centering
        \includegraphics[width=\linewidth]{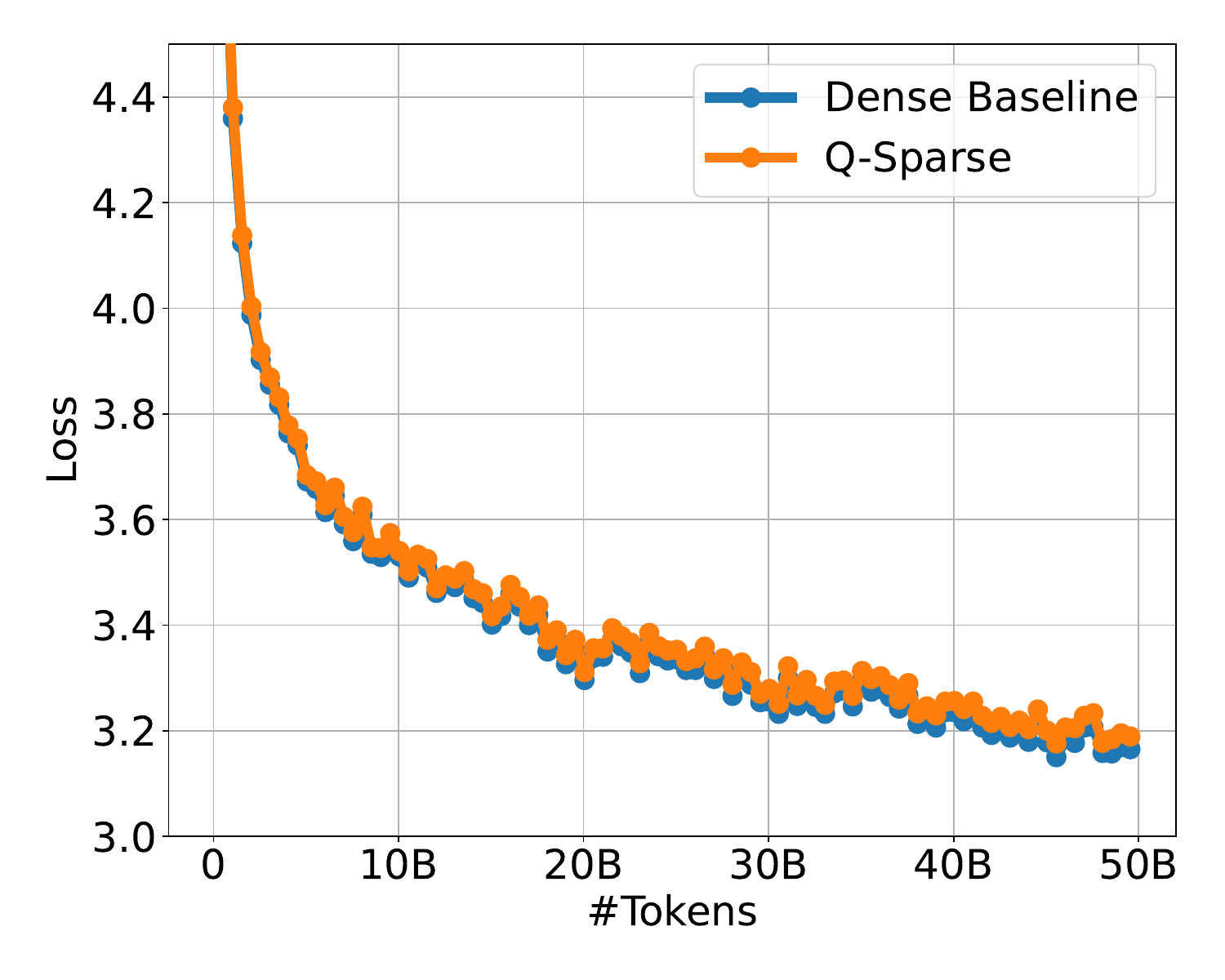}
        \caption{700M model size}
    \end{subfigure}
    \begin{subfigure}{0.48\textwidth}
        \centering
        \includegraphics[width=\linewidth]{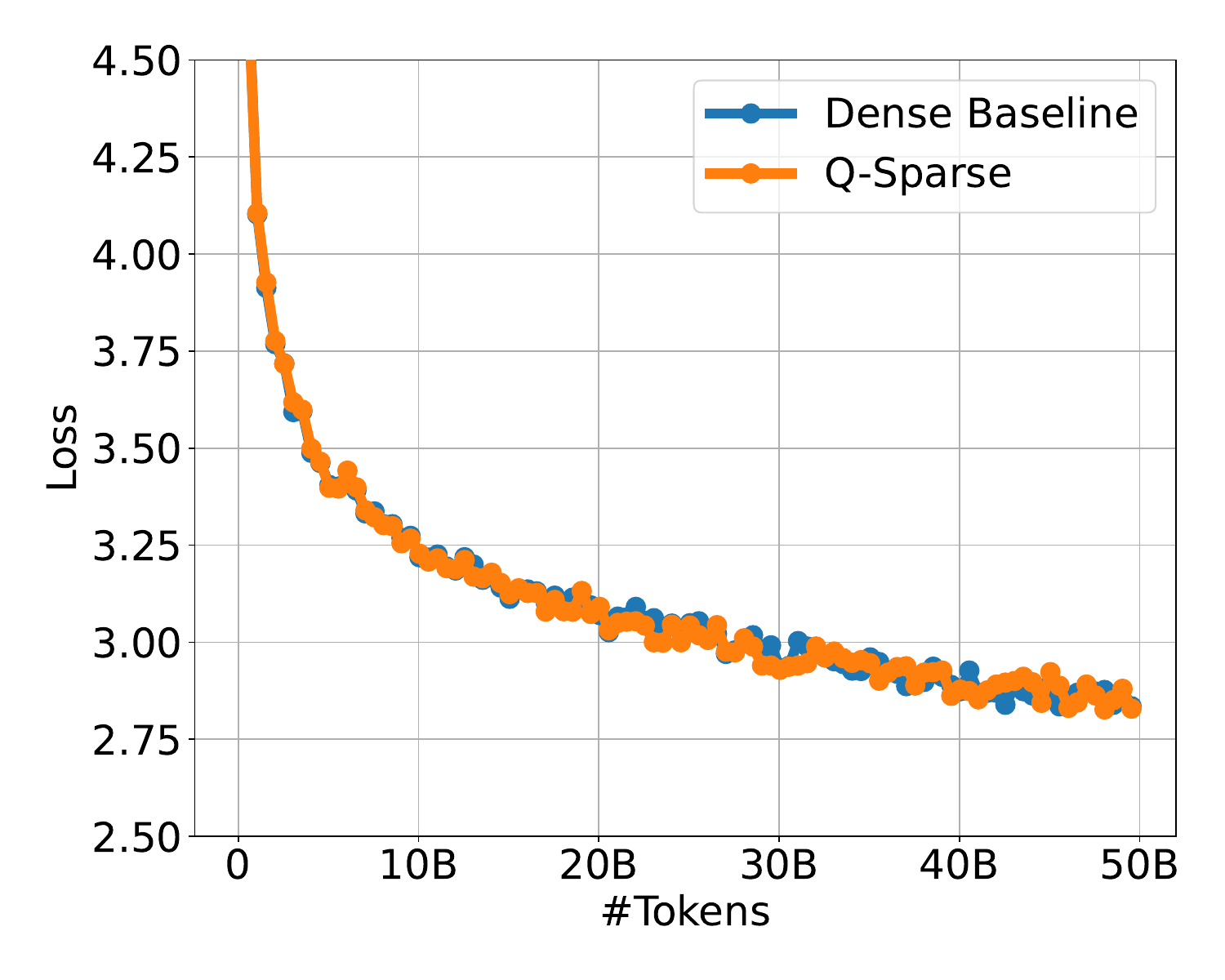}
        \caption{7B model size}
    \end{subfigure}
    \caption{The training loss curve of \our{} and the baseline with full-precision. We adopt top-$K$ as 70\% for \our{}, resulting in 40\% overall sparsity.}\label{fig:fp16_loss}
\end{figure}

\begin{figure}[t]
    \centering
    \begin{subfigure}{0.48\textwidth}
        \centering
        \includegraphics[width=\linewidth]{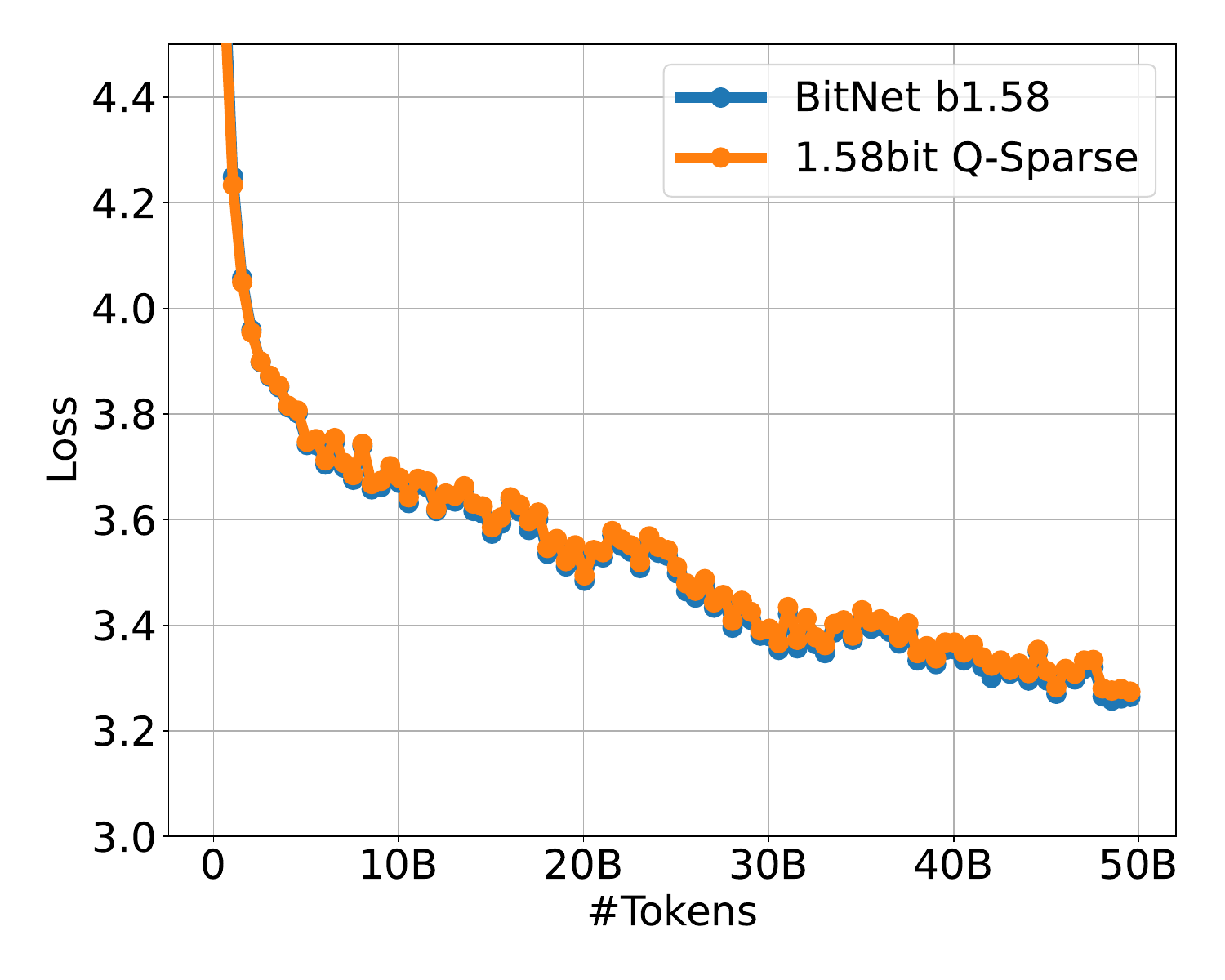}
        \caption{700M model size}
    \end{subfigure}
    \begin{subfigure}{0.48\textwidth}
        \centering
        \includegraphics[width=\linewidth]{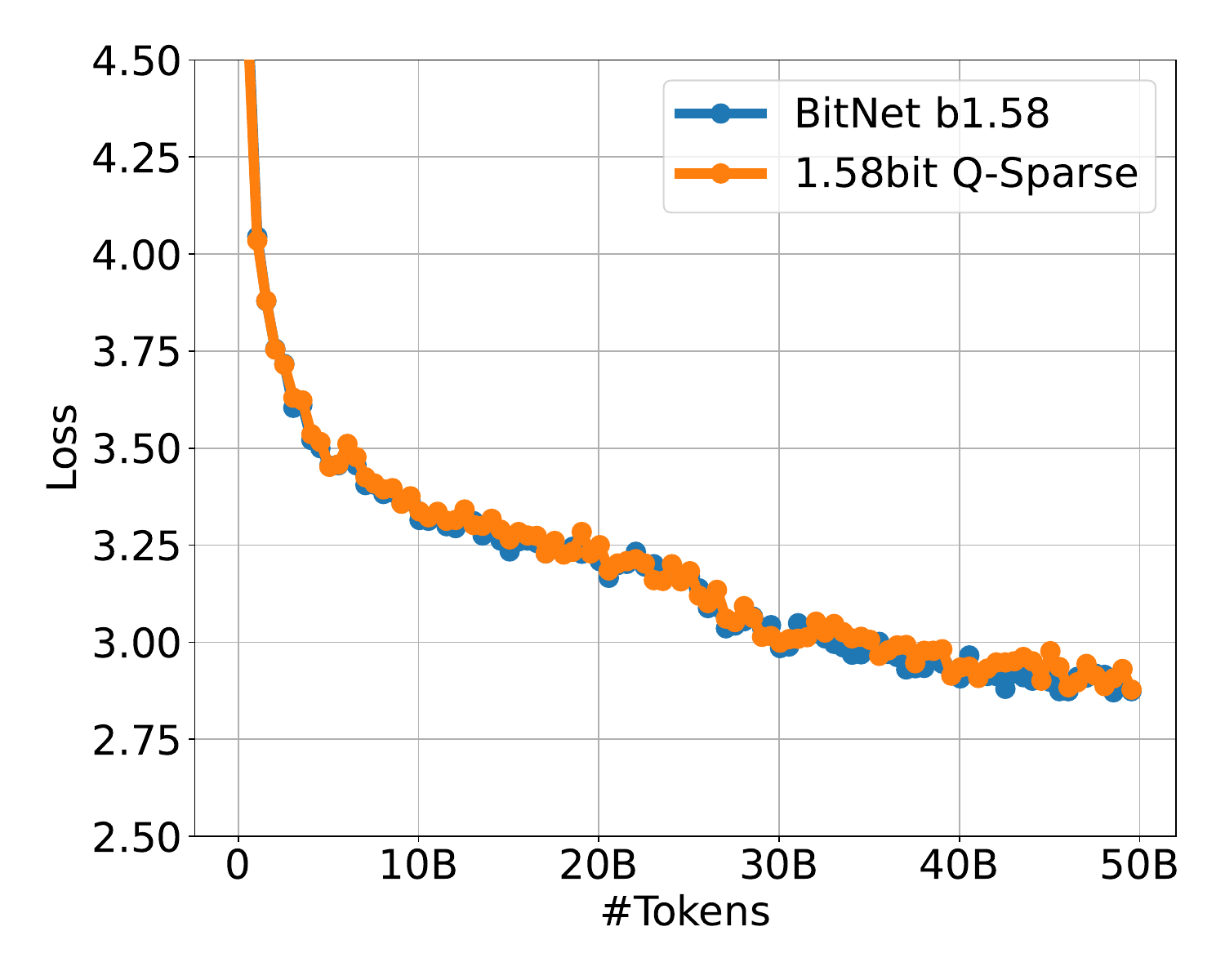}
        \caption{7B model size}
    \end{subfigure}
    \caption{The training loss curve of \our{} and the baseline with 1.58-bit weight. We adopt top-$K$ as 70\% for \our{}, resulting in 40\% overall sparsity.}\label{fig:bitnet_loss}
\end{figure}

\begin{figure}[t]
    \centering
    \begin{subfigure}{0.48\textwidth}
        \centering
        \includegraphics[width=\textwidth]{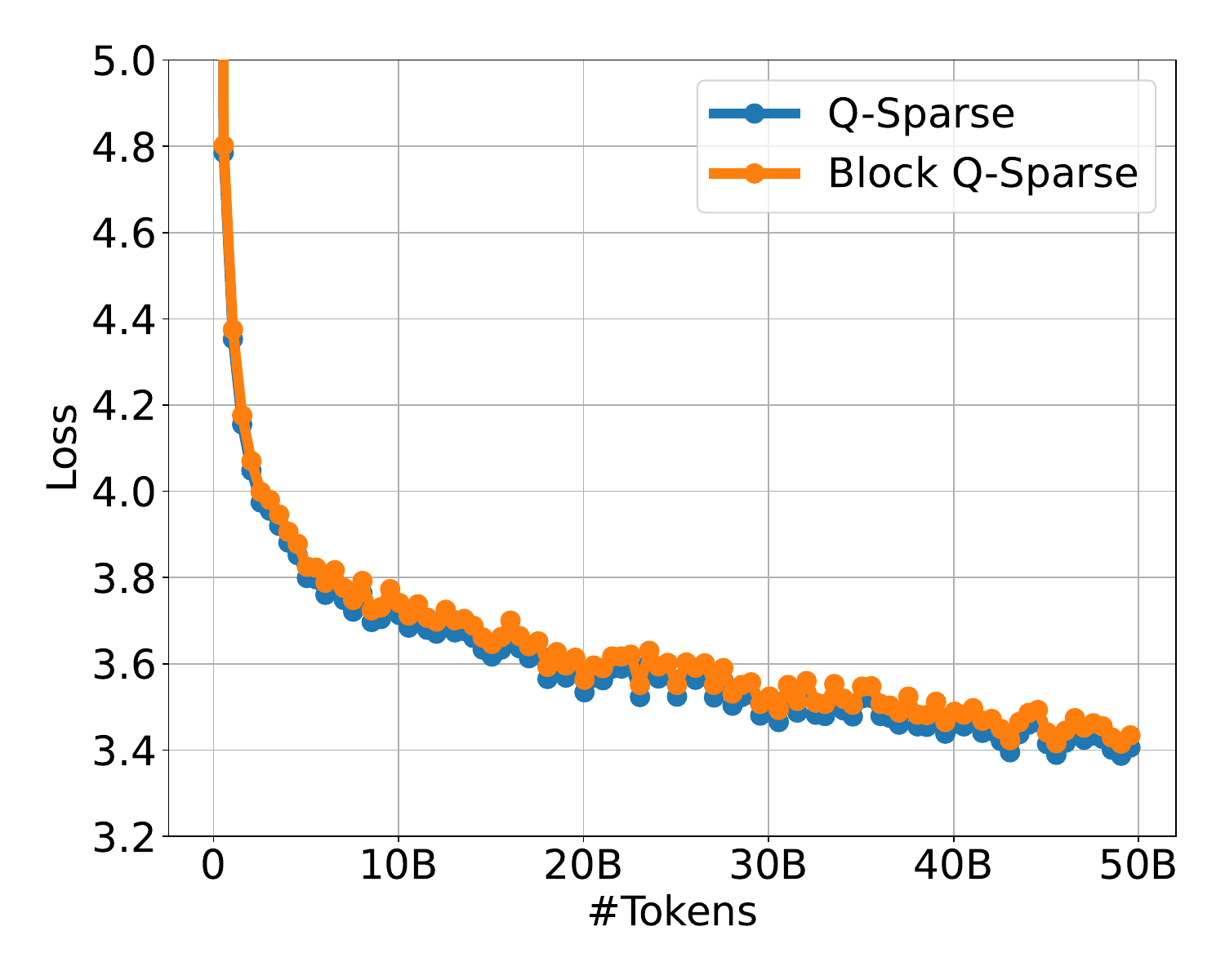}
        \caption{300M model size}
    \end{subfigure}
    \begin{subfigure}{0.48\textwidth}
        \centering
        \includegraphics[width=\textwidth]{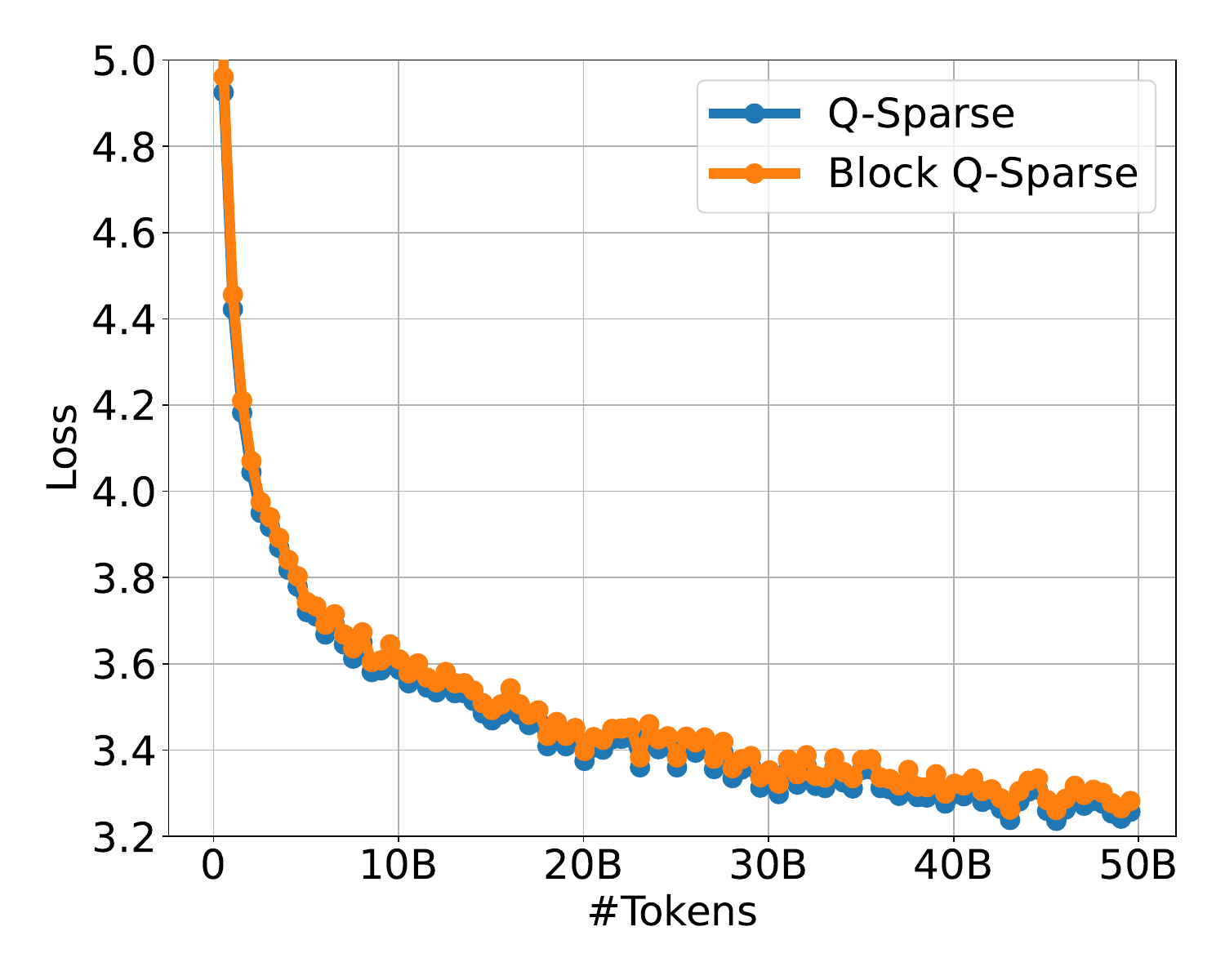}
        \caption{700M model size}
    \end{subfigure}
    \caption{The training loss curves for \our{} and Block \our{}. It shows that Block \our{} has a similar convergence to \our{} with the same sparsity.}\label{fig:block_sparse}
\end{figure}

\section{Experiments}

We conduct experiments to evaluate the effectiveness of \our{} in different settings, including training-from-scratch, continue-training of off-the-shelf LLMs, and finetuning.

\subsection{Training-from-Scratch}

\paragraph{Setting} We train a series of language models with \our{} in both full-precision and 1.58 bits. The models are trained with 50B tokens on the Redpajama dataset~\cite{redpajama}. We compare \our{} with the dense baselines with the same datasets and settings.

\paragraph{Results} The observed losses of the sparsely-activated models and the dense baselines are shown in Figure~\ref{fig:fp16_loss}. It shows that \our{} with 40\% sparsity ratio can match the performance of the dense baselines with the same model size and training tokens.

\paragraph{BitNet b1.58 + \our{}} We further evaluate the effectiveness of \our{} on 1-bit LLMs. We train a series of BitNet b1.58 models with \our{} of various scales. We plot the training loss curves of both \our{} and the BitNet b1.58 baseline. Figure~\ref{fig:bitnet_loss} shows that the performance of the sparsely-activated BitNet b1.58 models is better than the dense baselines with the same inference compute budget. It demonstrates that \our{} is compatible to 1-bit LLMs and their synergy can be used to optimize the performance of the models during inference.

\paragraph{Block \our{}} We evaluate the effectiveness of Block \our{}. We compare it with \our{} of the same sparsity ratio. 
The sparsity ratio is 50\%, and the block size is set to 32 (i.e., N:M=16:32). The experiments are performed with the model sizes of 300M and 700M. The training loss curves of \our{} and Block \our{} are shown in Figure~\ref{fig:block_sparse}. It shows that Block \our{} has a similar convergence to \our{} with the same sparsity. It demonstrates that Block \our{} can match the performance of \our{} when training from scratch.

\begin{figure}[t]
    \centering
    \begin{subfigure}{0.49\textwidth}
        \centering
        \includegraphics[width=\linewidth]{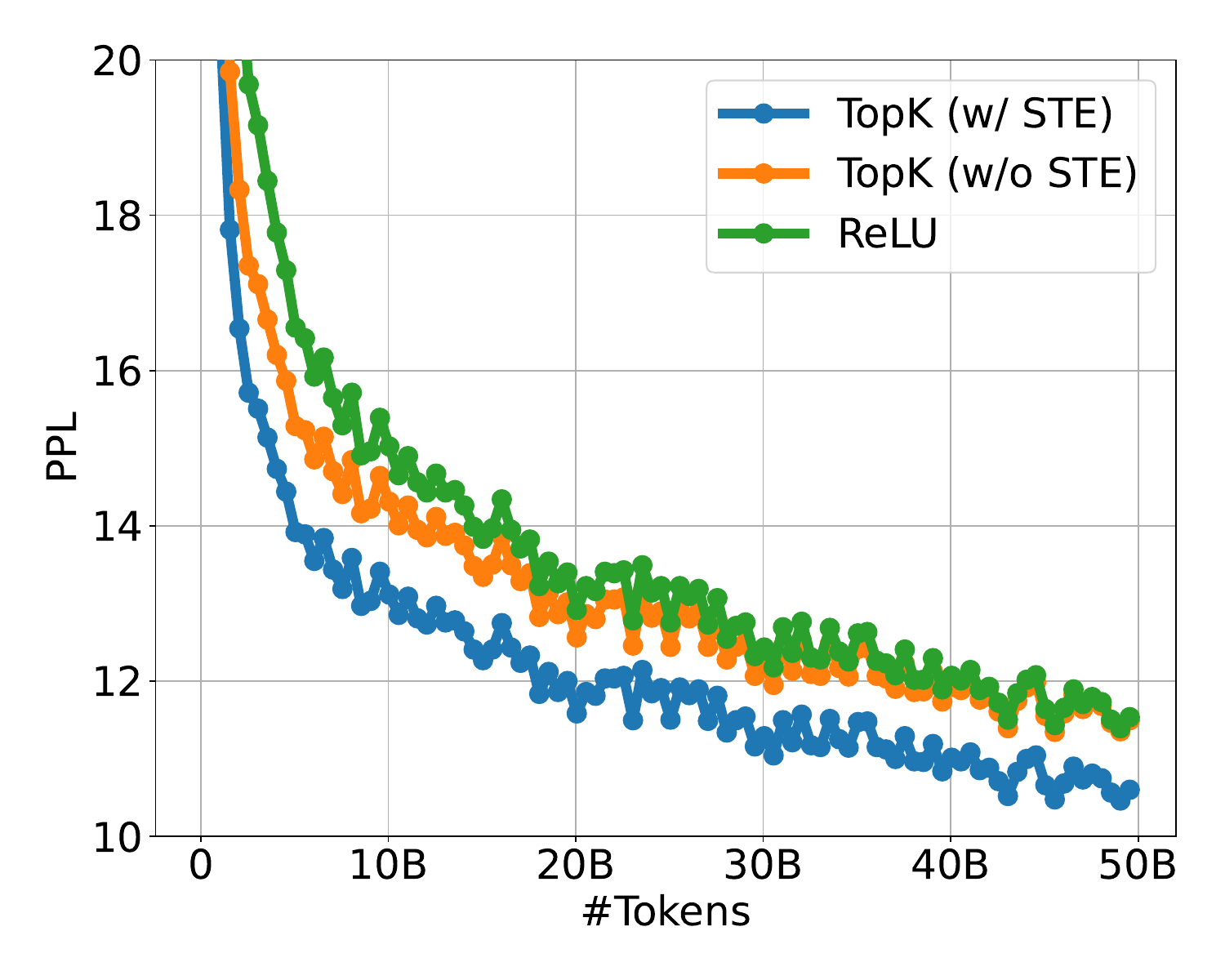}
    \end{subfigure}
    \begin{subfigure}{0.49\textwidth}
        \centering
        \includegraphics[width=\linewidth]{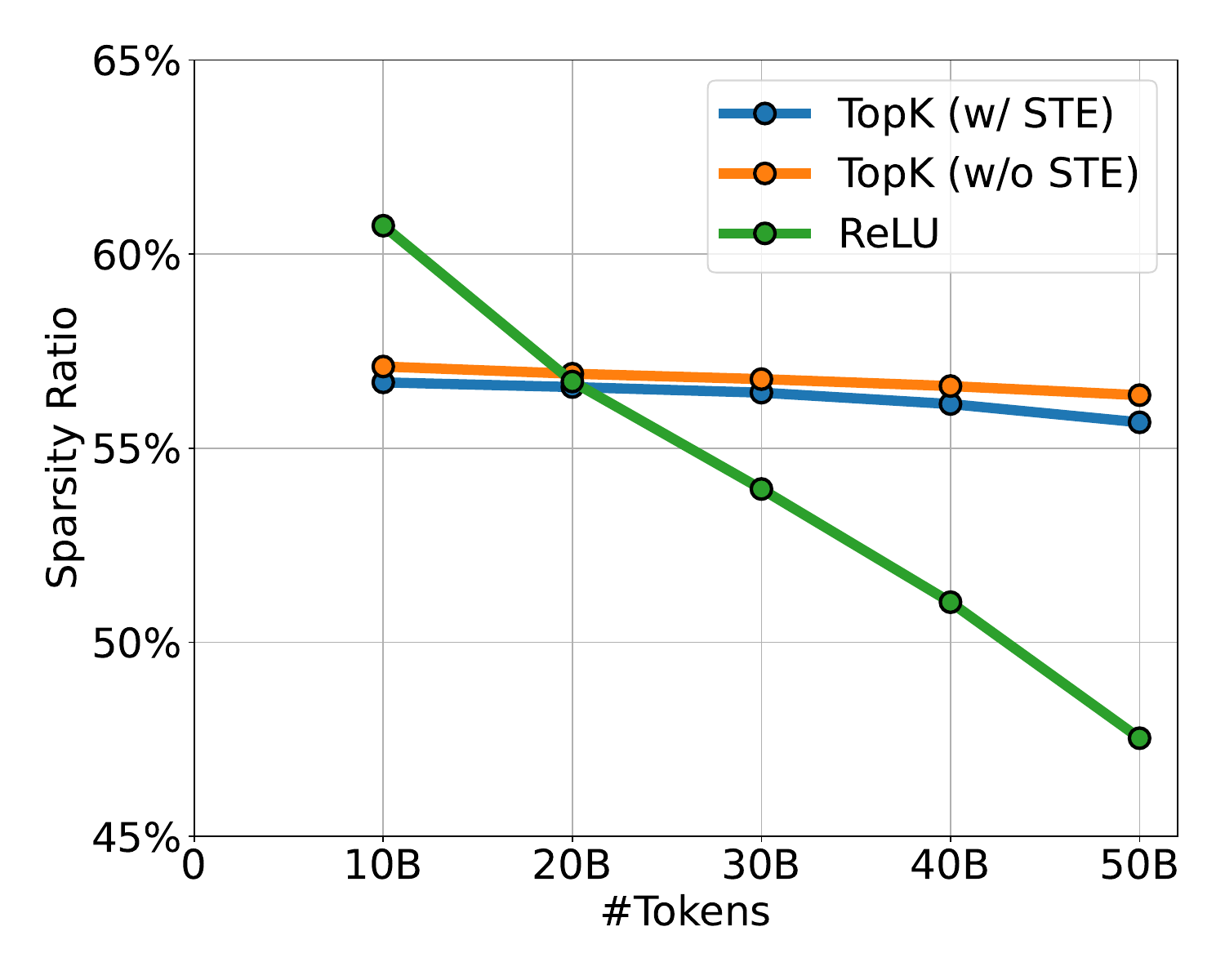}
    \end{subfigure}
    \caption{The training loss curves (Left) and the overall sparsity ratio (Right) of different sparsity functions. All models are trained with 300M size and 50B tokens.}\label{fig:ablation}
\end{figure}

\begin{figure}[t]
    \centering
    \begin{subfigure}{0.49\textwidth}
        \centering
        \includegraphics[width=\linewidth]{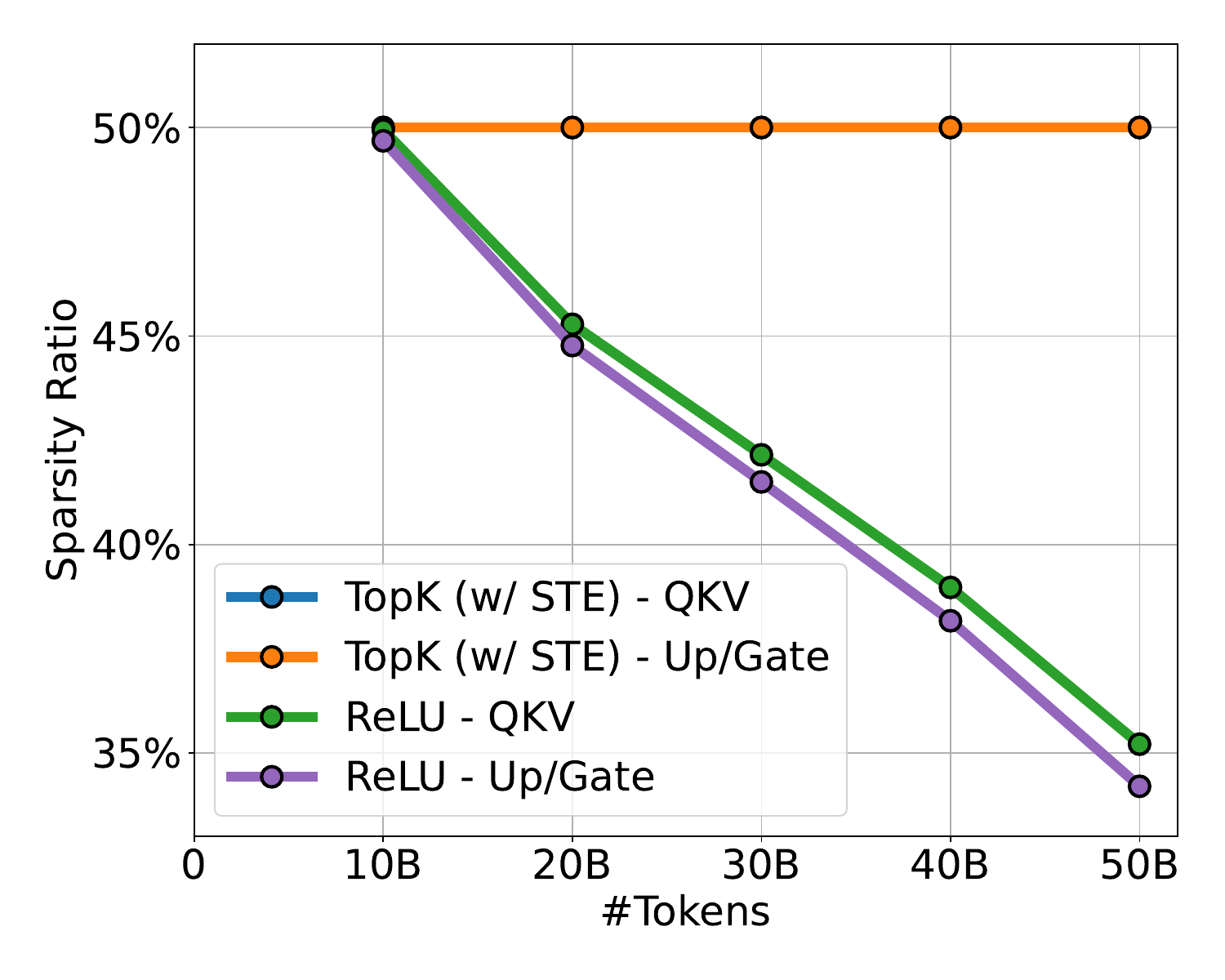}
    \end{subfigure}
    \begin{subfigure}{0.49\textwidth}
        \centering
        \includegraphics[width=\linewidth]{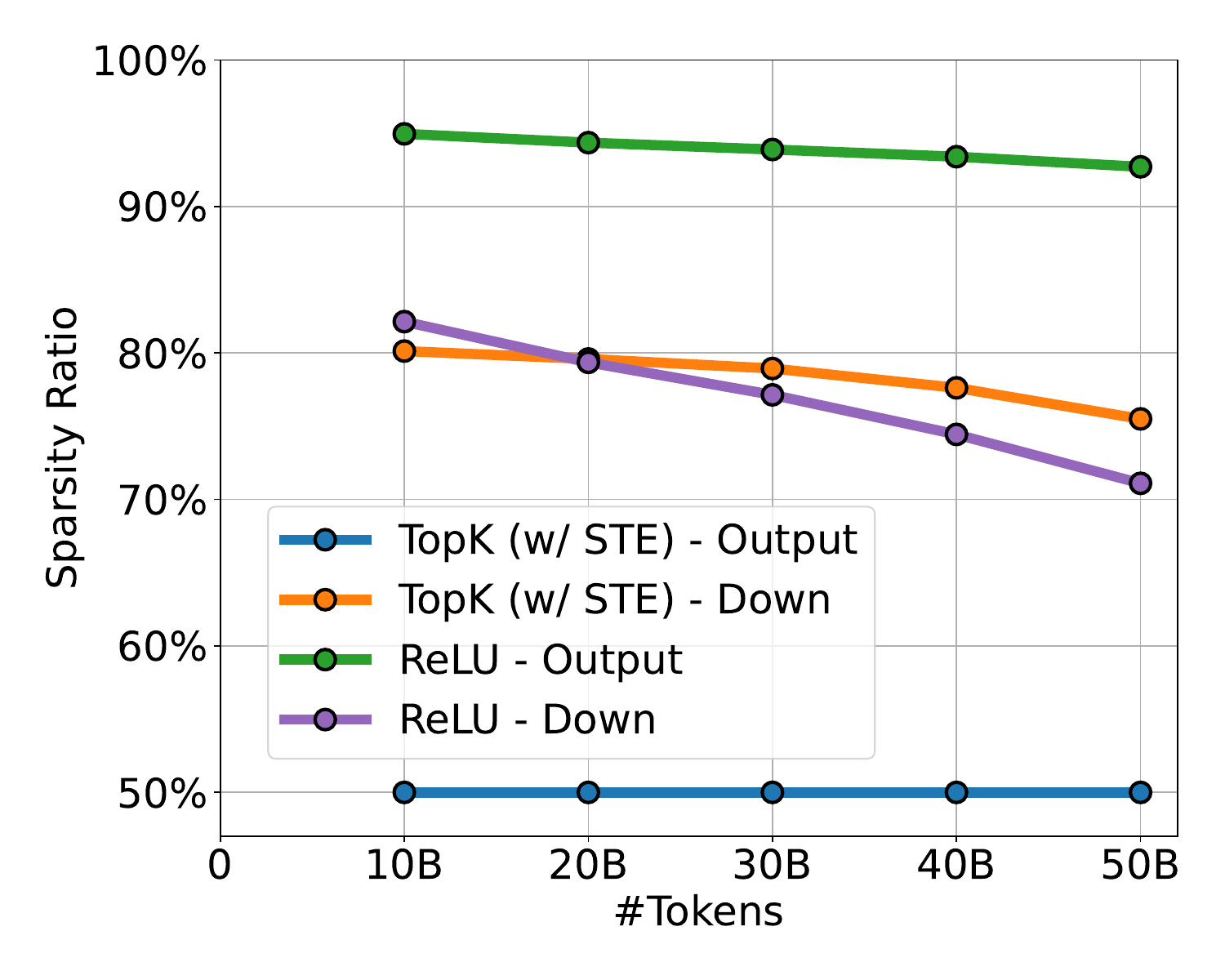}
    \end{subfigure}
    \caption{The sparsity ratio of each model's component of different sparsity functions.}\label{fig:ablation_detail}
\end{figure}

\paragraph{Ablation Study of top-K Sparisty and STE} To evaluate the effect of the top-K sparsity function, we compare the performance of the sparsely-activated models with the top-K sparsity function and the ReLU sparsity function. Moreover, we study the effect of the STE by comparing the models with and without STE. Figure~\ref{fig:ablation} illustrates the results. It shows that either removing STE or replacing with ReLU function significantly hurt the performance. Besides, the sparsity ratio of the models with the ReLU function decreases as the training processes. In constrast, the sparisty ratio remains unchanged with the top-K sparisty function. As shown in Figure~\ref{fig:ablation_detail}, we break down the contribution of the sparsity ratio from different components, finding that the decreasing sparisty is mainly from the QKV projection, the gating projection and the up projection of the feed-forward layers. This proves the superior of top-K over ReLU function.

\subsection{Continue-Training}

\paragraph{Setting} We continue-train the Mistral 7B model~\cite{qwen}  for 40B tokens on the FineWeb-Edu dataset~\cite{fineweb_edu}. We use the Sentencepiece tokenizer from Mistral to preprocess data. We use the batch size of 4M tokens and the learning rate of 5e-5. We use the Adam optimizer with the weight decay of 0.01. More training details can be found in Appendix~\ref{ap:hyper}.

\paragraph{Results} For a fair comparison, we continue-train the Mistral 7B model with the same recipe as the dense baseline. We compare \our{} with the ReLUfication~\cite{relu_back} and dReLU Sparsification~\cite{turbosparse} methods, which sparsify the model by changing the activation function. Following the origin paper~\cite{relu_back}, we adopt a two-stage training strategy that first replaces the non-ReLU activation and then adds the ReLU functions. For the dReLU Sparsification method, we implement the dReLU sparsification method following the origin paper~\cite{turbosparse}. We evaluate these models on a range of language tasks, including ARC-Challenge~\cite{arc}, HellaSwag~\cite{hellaswag}, Winogrande~\cite{winoGrande}, MMLU~\cite{mmlu} and TruthfulQA~\cite{truthfulqa}. Results are shown in Table~\ref{tab:fewshot}. It shows that \our{} achieves comparable performance to the dense baseline while being much more efficient at inference time. Moreover, \our{} outperforms the ReLUfication and dReLU Sparsification methods in terms of the performance and the sparsity ratio.

To break down the sparsity of each component in the model, we present the sparsity ratio of the query, key, value, output, up, down, and gate tensors in Table~\ref{tab:sparsity}. It shows that \our{} achieves a higher sparsity ratio than the ReLUfication and dReLU Sparsification methods. The sparsity ratio of the query, key, value, output, up, and down tensors is higher than 40\%, and the sparsity ratio of the gate tensor is higher than 60\%. It demonstrates that \our{} can achieve full sparsity of activations in LLMs.

\begin{table*}[t]
    \setlength{\tabcolsep}{6pt}
    \centering
    \begin{tabular}{lccccccc}
    \toprule
    \textbf{Models} & \textbf{Activated} & \textbf{ARC} & \textbf{HS} & \textbf{MMLU} & \textbf{WG} & \textbf{TQA} & \textbf{Avg.} \\
    \midrule
    Dense Baseline & 7.0B &  61.8 & 81.4	& 59.8	& 77.5	& 42.7	& 64.6 \\
    \midrule
    ReLUfication~\cite{relu_back} &  5.0B & 57.2 & 78.8	& 54.7	& 74.7 & 38.8 & 60.8 \\
    dReLU Sparsification~\cite{turbosparse} & 5.4B & 59.2 & 78.0	& 54.0	& 75.8	& 38.3	& 61.0 \\
    \midrule 
    \multirow{2}{*}{\textbf{\our{} (this work)}} & 2.9B & 59.0 	& 79.0 	& 55.6 & 74.0 	& 41.0 	& 61.7 \\
     & 3.8B & 60.5	& 80.7	& 58.0	& 75.9	& 43.5	& 63.7 \\
    \bottomrule
    \end{tabular}
    \caption{The results of the continue-training for \our{} and the baselines on the end tasks.}
    \label{tab:fewshot}
\end{table*}

\begin{table*}[t]
    \setlength{\tabcolsep}{6pt}
    \centering
    \begin{tabular}{lccccccc}
    \toprule
    \textbf{Models} & \textbf{Activated} & \textbf{QKV} & \textbf{Out} & \textbf{Up} & \textbf{Gate} & \textbf{Down} & \textbf{Overall} \\
    \midrule
    Dense Baseline & 7.0B & 0.0 & 0.0 & 0.0 & 0.0 & 0.0 & 0.0 \\
    \midrule
    ReLUfication~\cite{relu_back} & 5.0B &  12.3 & 0.0 & 10.3 & 10.3 & 79.3 & 28.3\\
    dReLU Sparsification~\cite{turbosparse} & 5.4B &  0.1 & 0.0 & 0.1 & 0.1 & 85.5 & 23.0 \\
    \midrule 
    \multirow{2}{*}{\textbf{\our{} (this work)}} & 2.9B &  51.4 & 50.0 & 50.0 & 50.0 & 80.0 & 58.2  \\
     & 3.8B & 42.0 & 40.0 & 40.0 & 40.0 & 60.4 & 45.7\\
    \bottomrule
    \end{tabular}
    \caption{The activated parameters and the sparsity ratio of the continue-training for \our{} and the baselines on the test set of Wikitext2.}
    \label{tab:sparsity}
\end{table*}

\subsection{Supervised Finetuning}
\label{sec:sft}

\begin{table*}[t]
    \setlength{\tabcolsep}{6pt}
    \centering
    \begin{tabular}{lccccccc}
    \toprule
    \textbf{Models} & \textbf{Activated} & \textbf{ARC} & \textbf{HS} & \textbf{MMLU} & \textbf{WG} & \textbf{TQA} & \textbf{Avg.} \\
    \midrule
    Qwen1.5-4B & 3.2B & 42.8 & 68.2 & 53.6 & 67.1 & 47.9 & 55.9 \\
    Qwen1.5-7B & 6.5B & 47.7 & 74.6 & 61.5 & 71.4 & 50.7 & 61.2 \\ 
    \midrule
    \multirow{2}{*}{\textbf{\our{}}} & 3.6B & 46.3 & 72.6 & 59.1 & 67.5 & 50.3 & 59.2  \\
     & 4.1B & 47.9 & 73.2 & 59.2 & 69.4 & 51.1 & 60.1  \\
     \midrule
    \midrule
    Mistral-7B & 7.0B & 62.5 & 82.6 & 61.2 & 77.6 & 50.3 & 66.8 \\
    \midrule
    \multirow{2}{*}{\textbf{\our{}}} &  3.8B & 60.5 & 81.5 & 60.0 &	77.1 & 50.5 & 65.9 \\
     & 4.3B &  61.4 & 81.6 &	60.6 & 77.6 & 50.7 & 66.4  \\
     \midrule
    \end{tabular}
    \caption{The results of the supervised fine-tuning for \our{} and the dense baselines on the end tasks.}
    \label{tab:sft}
\end{table*}

\begin{table*}[t]
    \setlength{\tabcolsep}{6pt}
    \centering
    \begin{tabular}{lccccccc}
    \toprule
    \textbf{Models} & \textbf{Activated} & \textbf{ARC} & \textbf{HS} & \textbf{MMLU} & \textbf{WG} & \textbf{TQA} & \textbf{Avg.} \\
    \midrule
    Qwen1.5-4B & 3.2B & 42.8 & 68.2 & 53.6 & 67.1 & 47.9 & 55.9 \\
    Qwen1.5-7B & 6.5B & 47.7 & 74.6 & 61.5 & 71.4 & 50.7 & 61.2 \\ 
    \midrule
    \multirow{2}{*}{\textbf{Block \our{}}} &  3.6B & 47.0 & 71.1 & 56.7 & 67.6 & 50.5 & 58.6 \\
     & 4.1B & 47.2 & 73.1 &	59.7 & 69.0 & 49.7 & 59.7  \\
    \midrule
    \midrule
    Mistral-7B & 7.0B & 62.5 & 82.6 & 61.2 & 77.6 & 50.3 & 66.8 \\
    \midrule
    \multirow{2}{*}{\textbf{Block \our{}}} &  3.8B & 59.7 & 80.6 & 58.7 &	75.5 & 50.3 & 65.0 \\
     & 4.3B &  60.0 & 81.4 &	59.9 & 76.8 & 51.3 & 65.9  \\
    \bottomrule
    \end{tabular}
    \caption{The results of the supervised fine-tuning for Block \our{} and the dense baselines on the end tasks.}
    \label{tab:block_sft}
\end{table*}

\paragraph{Setting} We finetune the base model of Mistral 7B~\cite{mistral} and Qwen1.5 7B~\cite{qwen} on Open-Orca dataset~\cite{OpenOrca} for both the dense baselines and \our{}. The batch size is set as 128. The learning rates are selected from \{3e-6, 5e-6, 7e-6\}. All models are trained with 1 epoch for a fair comparison. The hyper-parameters are detailed in Appendix~\ref{ap:hyper}. We conduct the evaluation for these models on a range of language tasks, including ARC-Challenge~\cite{arc}, HellaSwag~\cite{hellaswag}, Winogrande~\cite{winoGrande}, MMLU~\cite{mmlu} and TruthfulQA~\cite{truthfulqa}.

\paragraph{Results} The results are shown in Table~\ref{tab:sft}. It shows that \our{} with 3.6B activated parameters achieves significant better performance than the Qwen1.5 4B dense model. Moreover, \our{} with around 4B activated parameters achieves comparable performance to the Mistral 7B model and the Qwen1.5 7B model. It demonstrates that \our{} can be used to finetune a dense pretrained model to a much more efficient sparse model with almost no loss at accuracy.

\subsection{Evaluation of Block \our{}}

\paragraph{Setting} We finetune the base model of Mistral 7B~\cite{mistral} and Qwen1.5 7B~\cite{qwen} on Open-Orca dataset~\cite{OpenOrca} for Block \our{}. The block size is set as 32, which is recommended by the previous work~\cite{sparsenm} on N:M sparse kernels. The other hyper-parameters are consistent with the experiments shown in Section~\ref{sec:sft}.

\paragraph{Results} Table~\ref{tab:block_sft} summarizes the results for Block \our{}. Similar to the results of \our{}, Block \our{} achieves comparable performance to the dense baselines with much fewer activated parameters. It demonstrates that Block \our{} can be used for a much more efficient sparse model while supporting the batch mode.

\section{Discussion and Future Work}

\textbf{Scaling BitNet b1.58 + \our{} + YOCO}

We have shown promising results of combining 1-bit LLMs (i.e., BitNet b1.58) and fully sparse activations (i.e., \our{}). We are working on scaling up the training in terms of both model size and training tokens. Furthermore, we will incorporate YOCO~\cite{yoco} to address the issue of KV cache for LLM inference. The integration of BitNet, \our{}, and YOCO provides a comprehensive approach to optimizing all data types in LLM inference and deployment, which includes systematic optimization of model weights, activations, and KV cache.

\textbf{\our{} + MoE}

Mixture-of-Experts has been the most widely method to achieve sparse activations in LLMs. \our{} is orthogonal and can be seamlessly integrated with MoE.

%


\bibliography{bitnet}
\bibliographystyle{alpha}

\appendix

\section{Visualizations}
\label{ap:vis}

\begin{figure}[h]
    \centering
    \begin{subfigure}{0.49\textwidth}
        \centering
    \includegraphics[width=\textwidth]{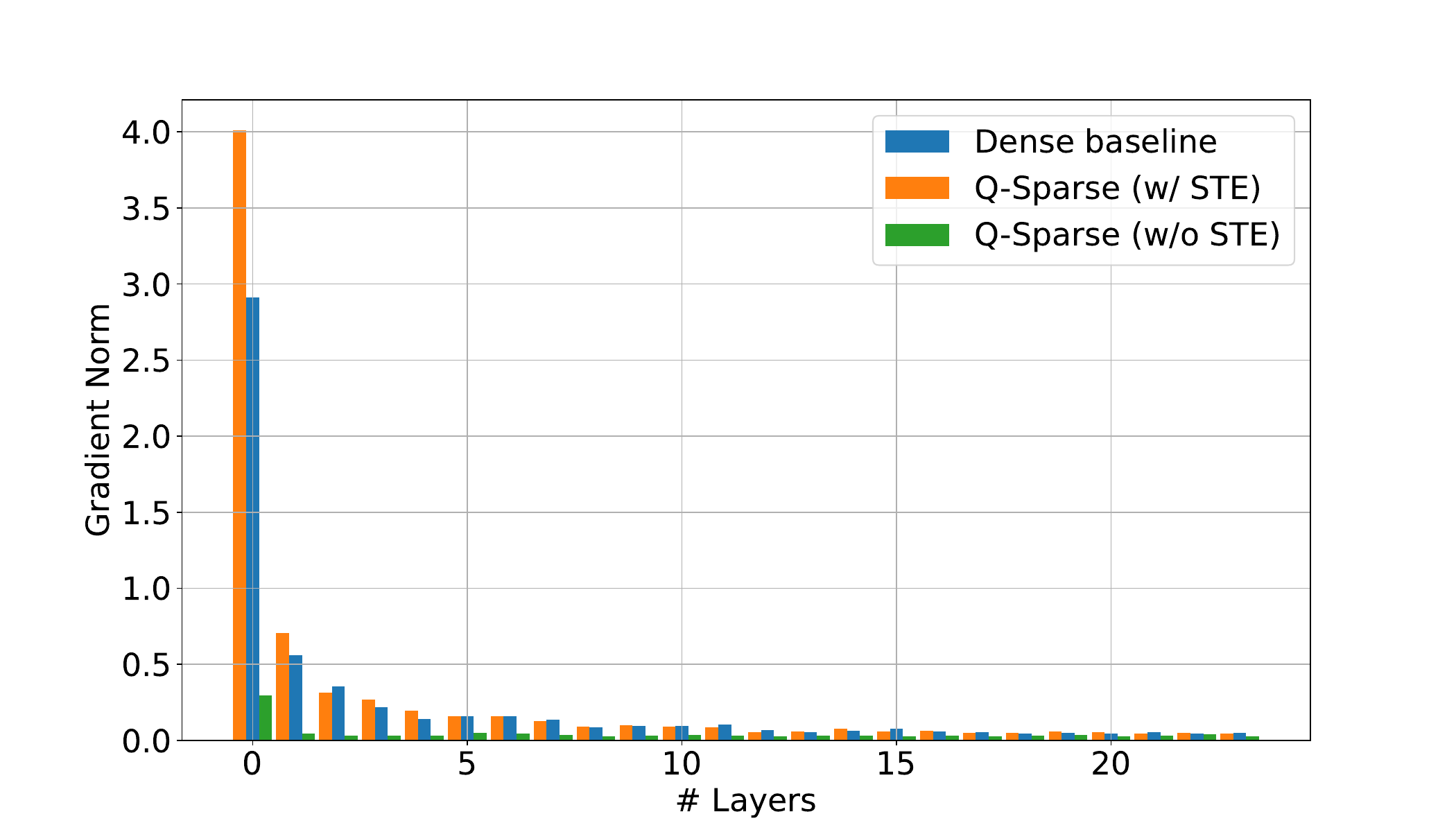}
        \caption{Query projection}
    \end{subfigure}
    \begin{subfigure}{0.49\textwidth}
        \centering
    \includegraphics[width=\textwidth]{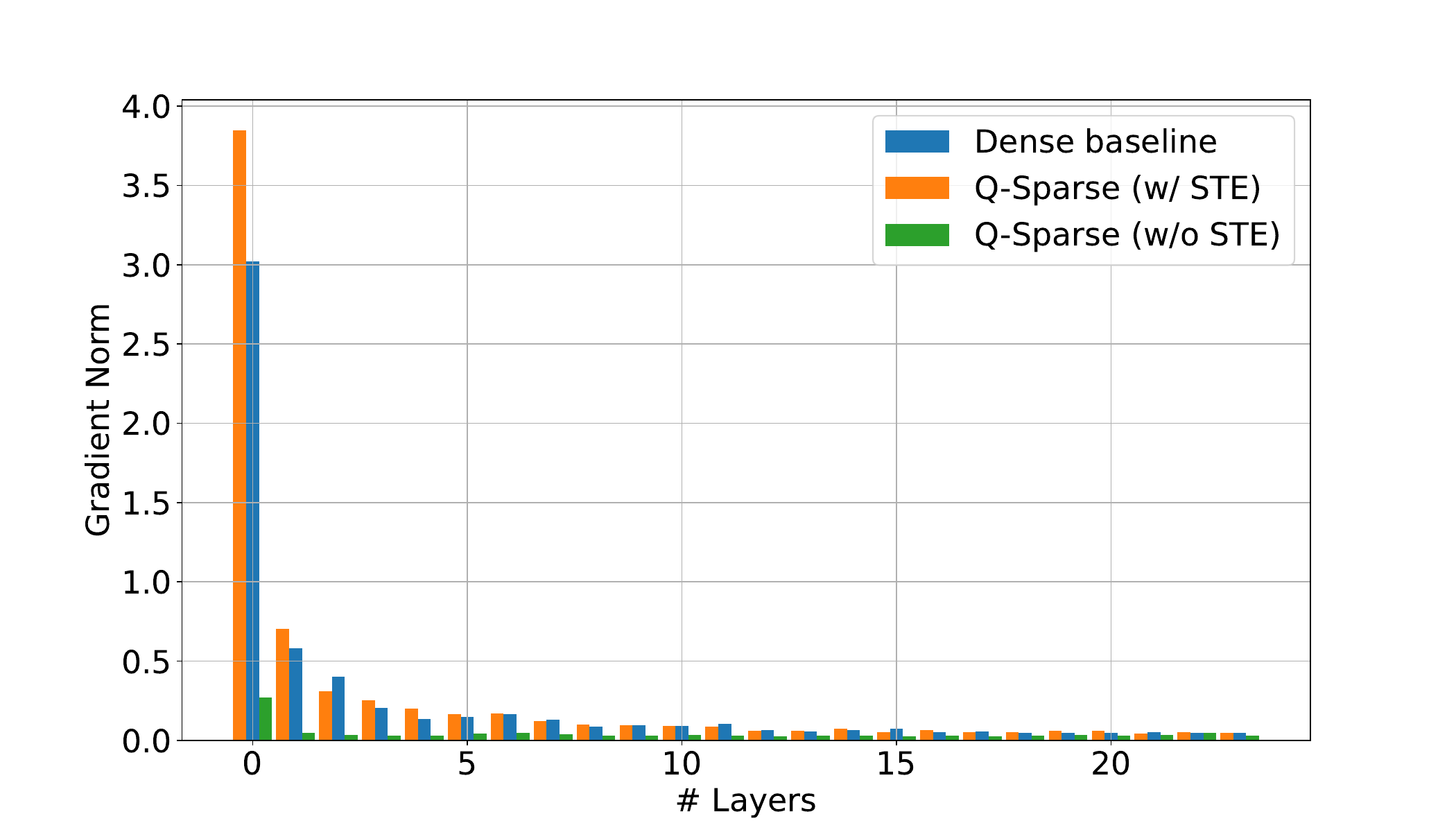}
    \caption{Key projection}
    \end{subfigure}
    \begin{subfigure}{0.49\textwidth}
        \centering
    \includegraphics[width=\textwidth]{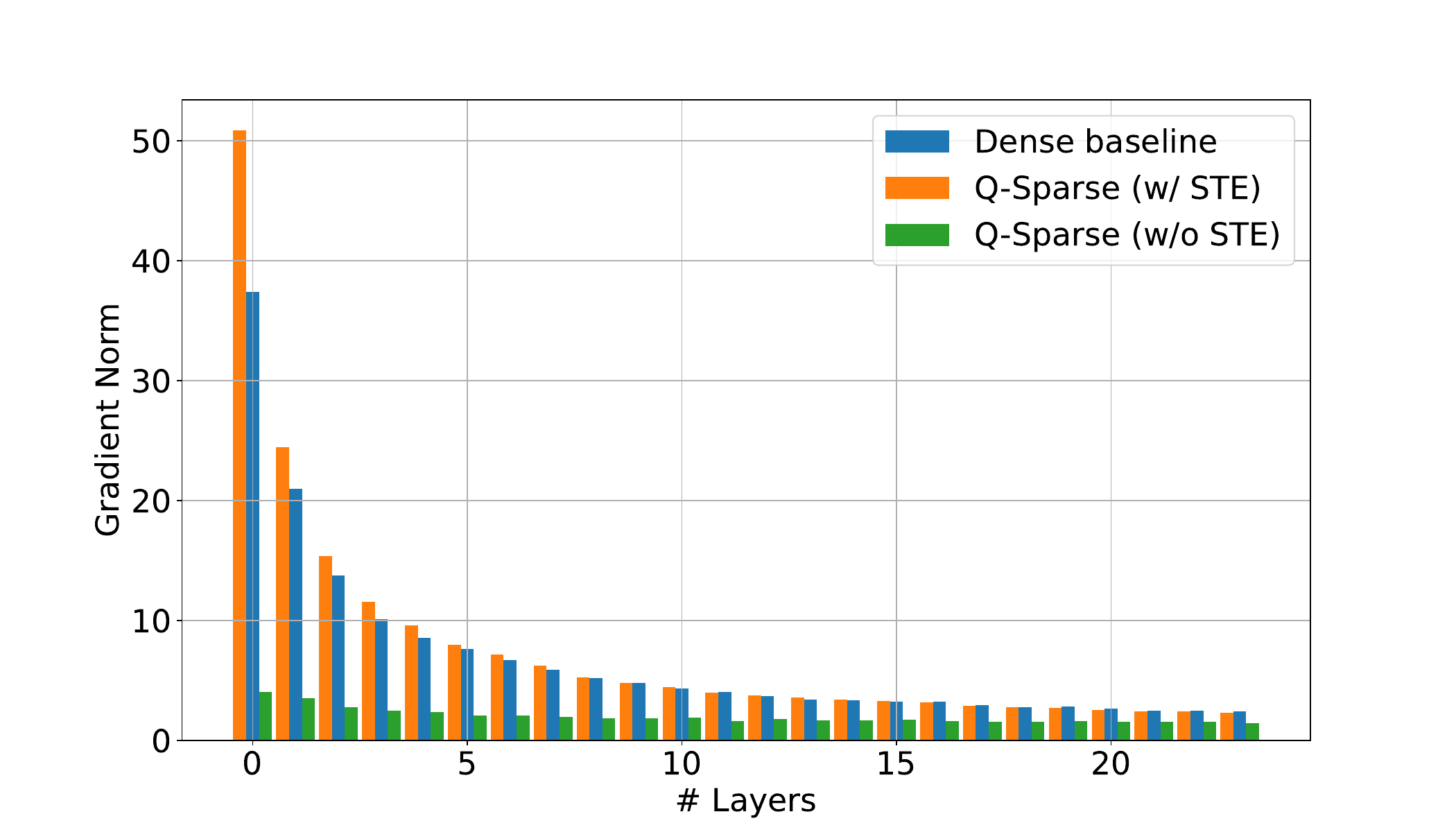}
    \caption{Value projection}
    \end{subfigure}
    \begin{subfigure}{0.49\textwidth}
        \centering
    \includegraphics[width=\textwidth]{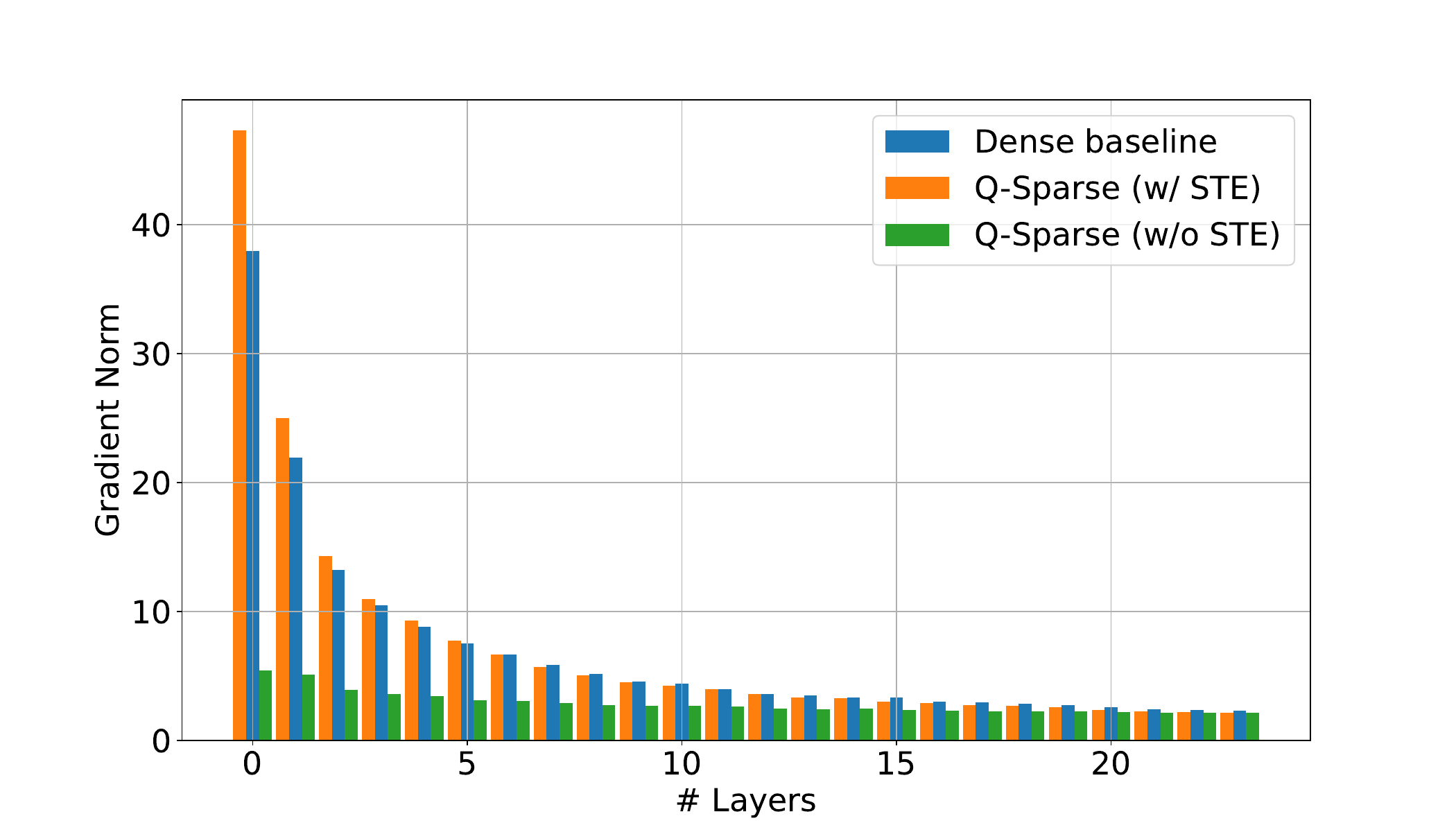}
    \caption{Output projection}
    \end{subfigure}
    \begin{subfigure}{0.32\textwidth}
        \centering
    \includegraphics[width=\textwidth]{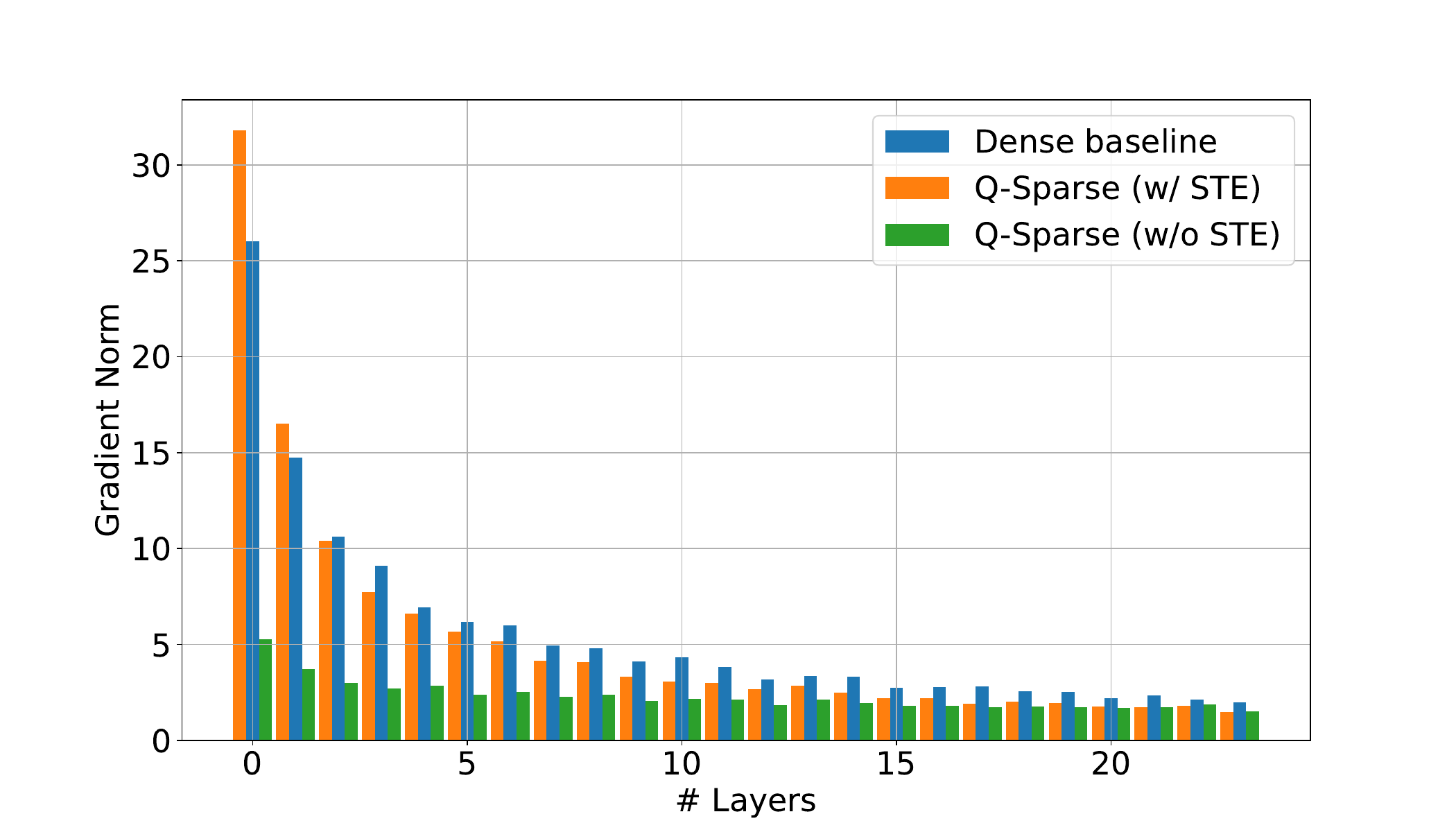}
    \caption{Gate projection}
    \end{subfigure}
    \begin{subfigure}{0.32\textwidth}
        \centering
    \includegraphics[width=\textwidth]{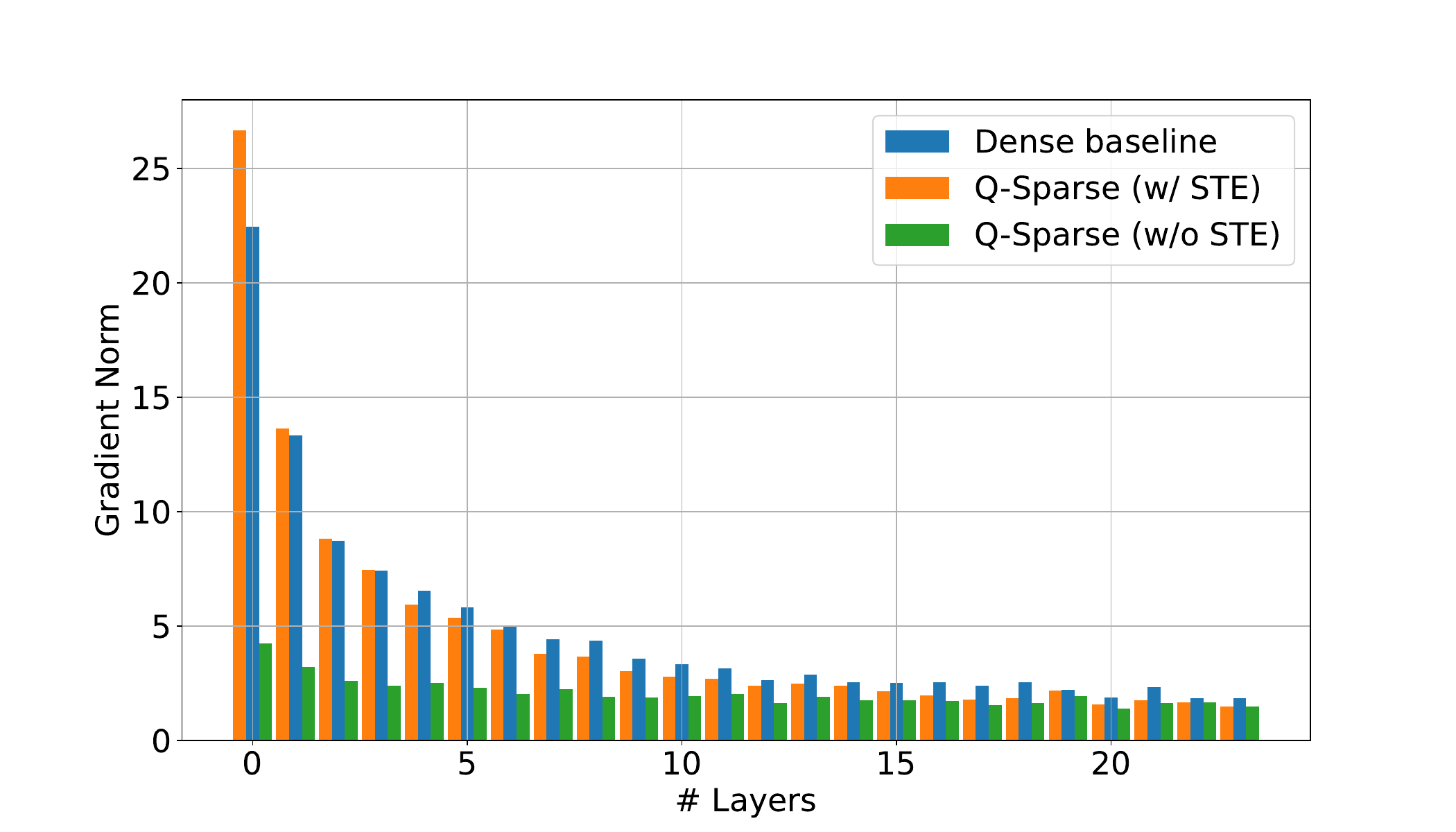}
    \caption{Up projection}
    \end{subfigure}
    \begin{subfigure}{0.32\textwidth}
        \centering
    \includegraphics[width=\textwidth]{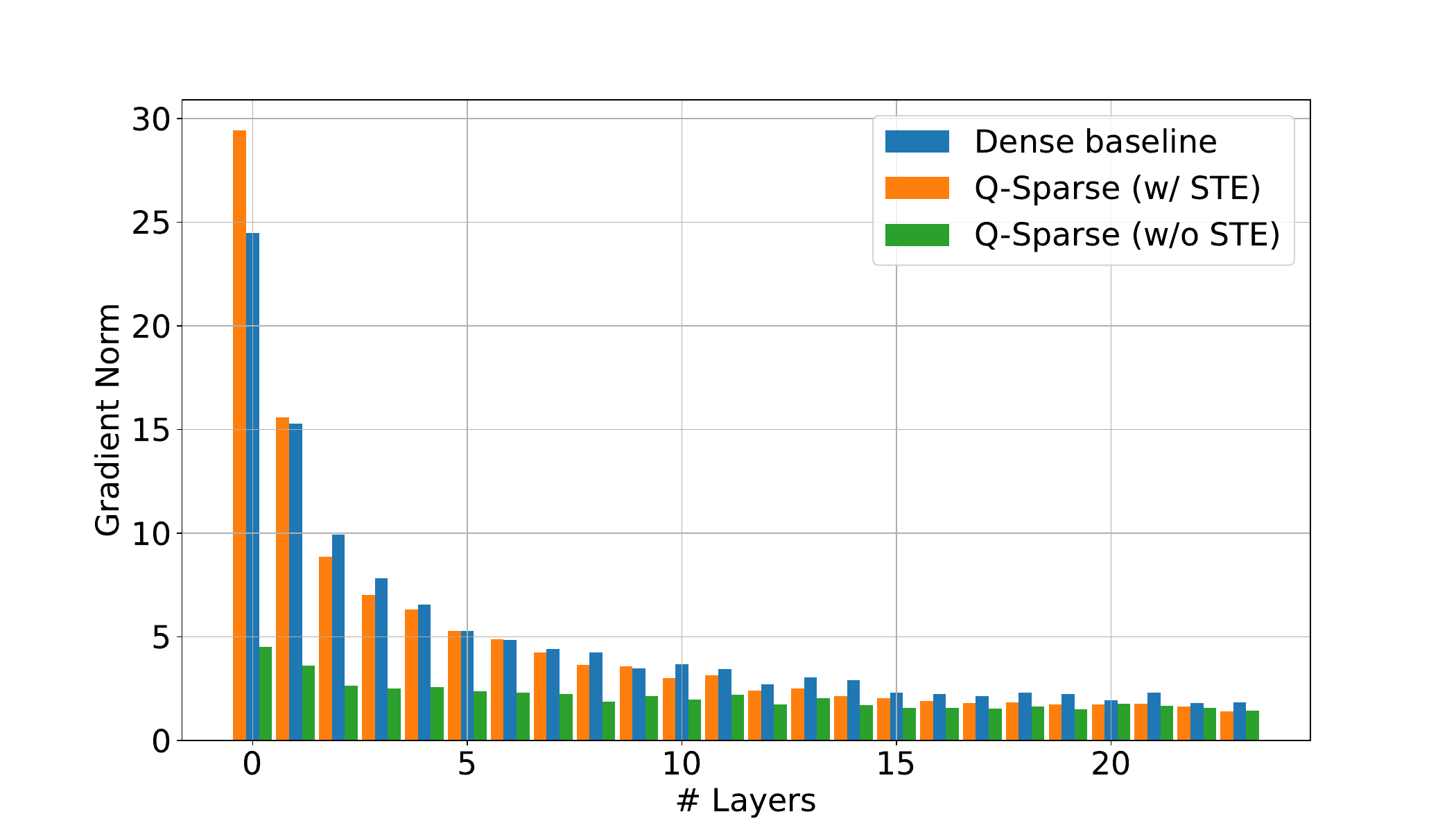}
    \caption{Down projection}
    \end{subfigure}
    \caption{The gradient magnitude of each linear projection of dense baseline, \our{} with and without STE estimator across different layers.}
    \label{fig:grad_gate}
\end{figure}

\section{Hyperparameters}
\label{ap:hyper}

\begin{table*}[h]
\setlength{\tabcolsep}{5pt}
\centering
\begin{tabular}{cccccccc}
\toprule
\bf Size & \bf Hidden Size & \bf GLU Size & \bf \#Heads & \bf \#Layers & \bf Seq Length \\
\midrule
    300M & 1024 & 2730 & 16 & 24 & 2048 \\
    700M & 1536 & 4096 & 24 & 24  & 2048 \\
    1.3B & 2048 & 5460 & 32 & 24   & 2048  \\
    7B & 4096 & 11008 & 32 & 32  & 2048  \\
\bottomrule
\end{tabular}
\caption{Model configurations for the scaling experiments of both BitNet b1.58 and \llama{} with \our{}.}
\label{tab:model_scaling}
\end{table*}

\begin{table*}[h]
    \centering
    \begin{tabular}{lllcccc}
    \toprule
    \bf Model & \bf Size & \bf Learning Rate & \bf Weight Decay & \bf Batch Size & \bf Adam $\beta$ \\
    \midrule
    \multirow{4}{*}{BitNet b1.58} & 300M &  $1.8\times10^{-3} \rightarrow 1.5\times10^{-3}$ & $0.1 \rightarrow 0$ & 0.5M & (0.9, 0.95) \\
    & 700M & $1.5\times10^{-3} \rightarrow 1\times10^{-3}$ & $0.1 \rightarrow 0$ & 0.5M & (0.9, 0.95) \\
    & 1.3B & $1.2\times10^{-3} \rightarrow 8\times10^{-4}$ & $0.1 \rightarrow 0$ & 0.5M & (0.9, 0.95)  \\
    & 7B &  $1\times10^{-3} \rightarrow 6\times10^{-4}$ & $0.1 \rightarrow 0$ & 0.5M & (0.9, 0.95) \\
    \midrule
    \multirow{4}{*}{\llama{}} & 300M & $6.0\times10^{-4}$ & 0.1 & 0.5M & (0.9, 0.95) \\
     & 700M & $2.5\times10^{-4}$ & 0.1 & 0.5M & (0.9, 0.95) \\
     & 1.3B & $2.0\times10^{-4}$ & 0.1 & 0.5M & (0.9, 0.95) \\
     & 7B &  $1.5\times10^{-4}$ & 0.1 & 0.5M & (0.9, 0.95) \\
    \bottomrule
    \end{tabular}
    \caption{Hyper-parameters for the scaling experiments of both BitNet b1.58 and \llama{} with \our{}.}
    \label{tab:optim_scaling}
\end{table*}

\begin{table*}[ht]
\centering
\begin{tabular}{l|c}
    \toprule
    \bf Hyperparameters & \bf Value  \\
    \midrule
    Training updates & 10K \\
    Tokens per sample & 4M \\
    Adam $\beta$ & (0.9, 0.95) \\
    Learning rate & 5e-5 \\
    End learning rate & 1e-6 \\
    Learning rate schedule & Polynomial decay \\
    Warmup updates & 375 \\
    \midrule
    Gradient clipping & 2.0 \\
    Dropout & \ding{55} \\
    Attention dropout & \ding{55} \\
    Weight decay & 0.01 \\
    \bottomrule
\end{tabular}
    \caption{
        Hyper-parameters for the continue-training of Mistral 7B with \our{} on Findweb Edu dataset.
    }
\label{tab:optim_cont_train}
\end{table*}

\begin{table*}[ht]
\centering
\begin{tabular}{l|c}
    \toprule
    \bf Hyperparameters & \bf Value  \\
    \midrule
    Training epoch & 1 \\
    Batch Size & 128 \\
    Adam $\beta$ & (0.9, 0.95) \\
    Learning rate & \{3e-6, 5e-6, 7e-6\} \\
    Learning rate schedule & Cosine decay \\
    Warmup ratio & 0.03 \\
    \midrule
    Dropout & \ding{55} \\
    Attention dropout & \ding{55} \\
    Weight decay & \ding{55} \\
    \bottomrule
\end{tabular}
    \caption{
        Hyper-parameters for the supervised fine-tuning of Mistral 7B and Qwen-1.5 7B with \our{} on OpenOrca dataset.
    }
\label{tab:optim_cont_train}
\end{table*}

\end{document}